\documentclass{article}

\usepackage{microtype}
\usepackage{graphicx}
\usepackage{booktabs} 

\usepackage{hyperref}

\usepackage[accepted]{icml2025}

\usepackage{amsmath}
\usepackage{amssymb}
\usepackage{mathtools}
\usepackage{amsthm}

\usepackage[capitalize,noabbrev]{cleveref}

\theoremstyle{plain}
\newtheorem{theorem}{Theorem}[section]
\newtheorem{proposition}[theorem]{Proposition}
\newtheorem{lemma}[theorem]{Lemma}

\theoremstyle{definition}
\newtheorem{definition}[theorem]{Definition}

\theoremstyle{remark}
\newtheorem{remark}[theorem]{Remark}

\usepackage[textsize=tiny]{todonotes}

\usepackage{subcaption}
\usepackage{epsf}
\usepackage{fancyhdr}
\usepackage{graphics}
\usepackage{psfrag}

\usepackage{wrapfig}
\usepackage{float}

\usepackage{algorithmic}

\usepackage{color}

\usepackage{bm,bbm}
\usepackage{multirow}
\usepackage{balance}

\usepackage{verbatim}

\makeatletter
\def\munderbar#1{\underline{\sbox\tw@{$#1$}\dp\tw@\z@\box\tw@}}
\makeatother

\newcommand{\be}{\begin{equation}}
\newcommand{\ee}{\end{equation}}
\newcommand{\bea}{\begin{equation*}\begin{aligned}}
\newcommand{\eea}{\end{aligned}\end{equation*}}

\newcommand{\R}{\mathbb{R}}
\newcommand{\N}{\mathbb{N}}

\newcommand{\G}{{\mathbb G}}

\newcommand{\calP}{{\mathcal P}}

\newcommand{\calS}{{\mathcal S}}
\newcommand{\calB}{{\mathcal B}}

\newcommand{\calW}{{\mathcal W}}

\newcommand{\calF}{{\mathcal F}}

\newcommand{\dd}{\mathrm{d}}

\newcommand{\Let}{\triangleq}

\newcommand{\norm}[1]{\left\lVert#1\right\rVert}

\icmltitlerunning{Scalable Sobolev IPM for Probability Measures on a Graph}

\begin{document}

\twocolumn[
\icmltitle{Scalable Sobolev IPM for Probability Measures on a Graph}



\icmlsetsymbol{equal}{*}

\begin{icmlauthorlist}
\icmlauthor{Tam Le}{equal,ism}
\icmlauthor{Truyen Nguyen}{equal,akron}
\icmlauthor{Hideitsu Hino}{ism}
\icmlauthor{Kenji Fukumizu}{ism}
\end{icmlauthorlist}

\icmlaffiliation{ism}{Department of Advanced Data Science, The Institute of Statistical Mathematics (ISM), Tokyo, Japan}
\icmlaffiliation{akron}{The University of Akron, Ohio, US}

\icmlcorrespondingauthor{Tam Le}{tam@ism.ac.jp}

\icmlkeywords{Sobolev IPM, probability measures on a graph, scalability, integral probability metrics (IPM), Sobolev norm, weighted Lp-norm}

\vskip 0.3in
]



\printAffiliationsAndNotice{\icmlEqualContribution} 

\begin{abstract}

We investigate the Sobolev IPM problem for probability measures supported on a graph metric space. Sobolev IPM is an important instance  of integral probability metrics (IPM), and is obtained by constraining a critic function within a unit ball defined by the Sobolev norm. In particular, it has been used to compare probability measures and is crucial for several theoretical works in machine learning. However, to our knowledge, there are no efficient algorithmic approaches to compute Sobolev IPM effectively, which hinders its practical applications. In this work, we establish a relation between Sobolev norm and weighted $L^p$-norm, and leverage it to propose a \emph{novel regularization} for Sobolev IPM. By exploiting the graph structure, we demonstrate that the regularized Sobolev IPM provides a \emph{closed-form} expression for fast computation. This advancement addresses long-standing computational challenges, and paves the way to apply Sobolev IPM for practical applications, even in large-scale settings. Additionally, the regularized Sobolev IPM is negative definite. Utilizing this property, we design positive-definite kernels upon the regularized Sobolev IPM, and provide preliminary evidences of their advantages for comparing probability measures on a given graph for document classification and topological data analysis.

\end{abstract}

\section{Introduction}
\label{sec:introduction}



Probability measures are widely used to represent objects of interest across various research fields. For instance, in natural language processing, documents can be viewed  as distributions over words~\citep{sparck1972statistical, kusner2015word, LYFC}, or distributions over topics~\citep{blei2003latent, yurochkin2019hierarchical}. 
In computer vision and graphics, 3D objects are often represented as point clouds, which are distributions of 3D data points~\citep{achlioptas2018learning, hua2018pointwise, wang2019dynamic, wu2019pointconv, nguyen2021point}. In topological data analysis (TDA), persistence diagrams (PD) are used to represent complex structural objects~\citep{le2018persistence, rieck2019persistent, zhao2019learning, divol2021estimation}.  Specifically, PDs can be regarded as distributions of 2D data points, where each point characterizes the birth and death time of a particular topological feature~\citep{edelsbrunner2008persistent}.



Integral probability metrics (IPMs) are a powerful class of distances for comparing probability measures~\citep{muller1997integral}. Essentially, IPMs identify a critic function (or witness function) that maximizes the discrimination between data points sampled from two input probability measures. IPMs have been applied in numerous theoretical studies and practical applications~\citep{sriperumbudur2009integral, gretton2012kernel, peyre2019computational, liang2019estimating, uppal2019nonparametric, uppal2020robust, nadjahi2020statistical, Kolouri2020Sliced}.

In this work, we study Sobolev IPM problem for probability measures supported on a graph. 
Sobolev IPM is an important instance of IPM, and is obtained by constraining the critic function within a unit ball defined by the Sobolev norm~\citep{adams2003sobolev}. 
Sobolev IPM plays a crucial role in several theoretical works, such as analyzing convergence rates of density estimation with generative adversarial networks (GANs) and studying error bounds for deep ReLU discriminator networks in GANs~\citep{liang2017well_arXiv, liang2021well_jmlr, singh2018nonparametric}. 
However, to our knowledge, a long-standing challenge for Sobolev IPM is that there are no efficient algorithmic approaches to compute it effectively which limits its applications in practice. 
To address this issue, we propose a \emph{novel regularization} for the Sobolev IPM problem. 
By leveraging the graph structure, we demonstrate that the regularized Sobolev IPM yields a \emph{closed-form} expression for fast computation,
thereby paving the way for its application in various domains, even for large-scale settings.

\paragraph{Contribution.} Our contributions are three-fold as follows:

\begin{itemize}
\item We propose a \emph{novel regularization} for Sobolev IPM that provides a closed-form expression for fast computation, overcoming the long-standing computational challenge and facilitating its application, particularly in large-scale settings.

\item We prove that the regularized Sobolev IPM is a metric and  show its \emph{equivalence} to the original Sobolev IPM. Additionally, we establish its connections to Sobolev transport (ST) which is a scalable variant of optimal transport (OT) for measures on a graph, and also its connection to the traditional OT on a graph.

\item We demonstrate that the regularized Sobolev IPM is negative definite. We then design positive definite kernels based on the regularized Sobolev IPM and provide preliminary evidence of their advantages to compare probability measures on a given graph in document classification and topological data analysis (TDA).
\end{itemize}

\paragraph{Organization.} In \S\ref{sec:preliminaries}, we introduce the notations used throughout our proposals. 
We detail our novel regularization for Sobolev IPM in~\S\ref{sec:regularized_SobolevIPM}. In \S\ref{sec:properties_regSobolevIPM}, we prove the metric property for the regularized Sobolev IPM and establish its connection to the original Sobolev IPM, ST, and OT on a graph. Additionally, we demonstrate the negative definiteness for the regularized Sobolev IPM and propose positive definite kernels based on it. In~\S\ref{sec:related_works}, we discuss related works. In \S\ref{sec:experiments}, we empirically illustrate the computational advantages of the regularized Sobolev IPM and provide preliminary evidence of its benefits in document classification and TDA. Finally, in \S\ref{sec:conclusion}, we offer concluding remarks. Proofs for theoretical results and additional materials are deferred to the Appendices. Additionally, we have released code for our proposed approach.\footnote{The code repository is on \url{https://github.com/lttam/Sobolev-IPM}.}

\section{Preliminaries}
\label{sec:preliminaries}

In this section, we briefly review the graph setting for probability measures. We also introduce some notations and functional spaces on the graph.


\begin{figure}
  \begin{center}
    \includegraphics[width=0.23\textwidth]{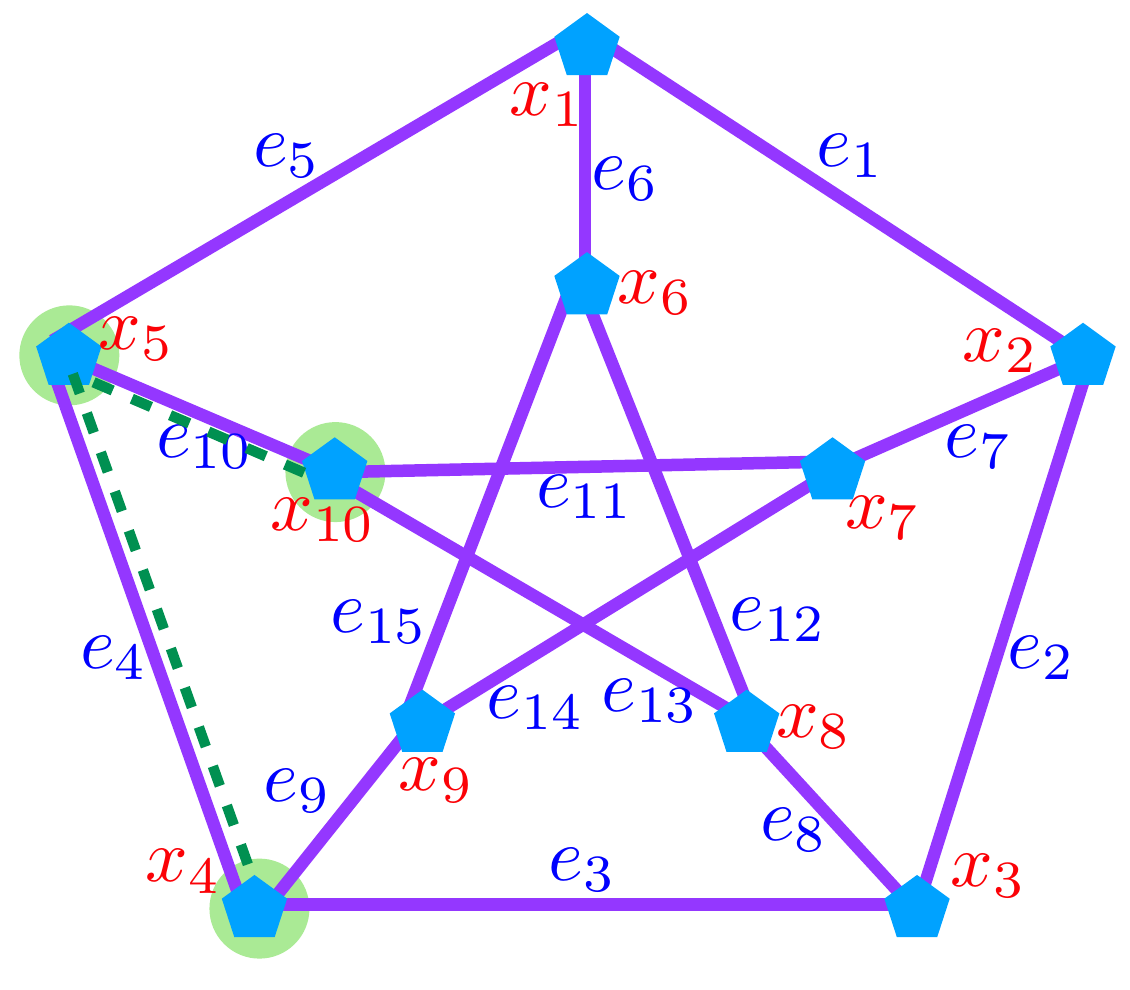}
  \end{center}
  \vspace{-6pt}
  \caption{A geodetic graph with $10$ nodes $\left\{x_1, x_2, \dotsc, x_{10}\right\}$ and $15$ edges $\left\{e_1, e_2, \dotsc, e_{15}\right\}$, and each edge length equals to $1$, i.e., $w_{e_j}=1, \forall j$. For any $x_i, x_j$, there is a unique shortest path between them, with a length $2$. Therefore, it satisfies the uniqueness property of the shortest paths. Let $x_1$ be the unique-path root node (i.e., $z_0 = x_1$), and subgraph $\widetilde{\G}$ containing $3$ nodes $\left\{x_4, x_5, x_{10}\right\}$ and $2$ edges $\left\{e_{4}, e_{10}\right\}$, then we have $\Lambda(x_5) = \gamma(e_5) = \widetilde{\G}$.}
  \label{fg:GeodeticGraph}
 \vspace{-6pt}
\end{figure}

\paragraph{Graph.} We follow the same graph setting in~\citet{le2022st} for measures and functions. Precisely, let $V, E$ 
 denote the sets of nodes and edges respectively. We consider a connected, undirected, and physical\footnote{In the sense that $V$ is a subset of Euclidean space $\R^n$; each edge $e \in E$ is the standard line segment in $\R^n$ connecting the two vertices  of the edge $e$.} graph $\G = (V,E)$ with positive edge lengths $\{w_e\}_{e\in E}$. The continuous graph setting is adopted, i.e. 
 $\G$ is regarded as the collection of all nodes in $V$ together with all points forming the edges in $E$. Additionally, $\G$ is equipped with the graph metric $d_{\G}(x,y)$, which is defined as the length of the shortest path connecting $x$ and $y$ in $\G$. We assume that there exists a fixed root node $z_0 \in V$ such that the shortest path between $z_0$ and $x$ is unique for any $x \in \G$, i.e., the uniqueness property of the shortest paths~\citep{le2022st}. \footnote{See Figure~\ref{fg:GeodeticGraph} for an illustration of a geodetic graph which satisfies the uniqueness property of the shortest paths.} 

Denote $[x, z]$ as the shortest path between $x$ and $z$ in $\G$. Then for given $x \in \G$ and edge $e \in E$, we define the sets $\Lambda(x)$ and $\gamma_e$ as follows
\begin{eqnarray}\label{sub-graph}
 && \Lambda(x) := \big\{y\in \G: \, x\in [z_0,y]\big\}, \nonumber \\
 && \gamma_e := \big\{y\in \G: \, e\subseteq  [z_0,y]\big\}.
\end{eqnarray}
See Figure~\ref{fg:GeodeticGraph} for an illustration of these graph notions.

\paragraph{Measures and functions on graph.} Let $\calP(\G)$ (resp. $\calP(\G \times\G)$) denote the set of all nonnegative Borel measures on $\G$ (resp. $\G\times\G$) with a finite mass.

By a continuous function $f$ on $\G$, we mean that  $f: \G\to \R$ is continuous w.r.t.~the topology on $\G$ induced by the Euclidean distance. Henceforth, $C(\G)$ stands for the set of all continuous functions on $\G$. Similar notation is used for continuous functions on $\G \times \G$. 

For a nonnegative Borel measure $\lambda$ on $\G$ and $1 \le p < \infty$, let $L^p( \G, \lambda)$ denote the space of all Borel measurable functions $ f:\G\to \R$ such that $\int_\G |f(x)|^p \lambda(\mathrm{d}x) <\infty$. Then $L^p( \G, \lambda)$ is a normed space with the norm being defined by 
\[
\|f\|_{L^p} := \left(\int_\G |f(x)|^p \lambda(\dd x)\right)^\frac{1}{p}.
\]
Additionally, let $\hat{w}$ be a positive weight function on $\G$, i.e., $\hat{w}(x) > 0$ for every $x \in \G$. Consider the weighted $L_{\hat w}^p( \G, \lambda)$  as the space of all Borel measurable functions $f:\G\to \R$ such that $\int_\G \hat{w}(x)|f(x)|^p \lambda(\mathrm{d}x) <\infty$. It is a normed space with the following  weighted $L^p$-norm 
\[
\|f\|_{L_{\hat w}^p} := \left(\int_\G \hat{w}(x) |f(x)|^p \lambda(\dd x)\right)^\frac{1}{p}.
\]
The spaces $L^p( \G, \lambda)$, $L_{\hat w}^p( \G, \lambda)$ 
 and their corresponding norms can also be defined for the case $p=\infty$. Further details are placed in Appendix \S\ref{app:sec:FuncSpaces}.



\section{Regularized Sobolev IPM}
\label{sec:regularized_SobolevIPM}

In this section, we consider the Sobolev IPM problem for probability measures on a graph. We show that the Sobolev norm  of $f$ is equivalent to a weighted $L^p$-norm of $f'$ with a specific weight function. Then 
we propose to leverage the weighted $L^p$-norm to regularize the Sobolev IPM. It is also demonstrated that  a closed-form expression can be derived for the regularized Sobolev IPM.

\subsection{Sobolev IPM}

We first introduce the graph-based Sobolev space~\citep{le2022st} and its Sobolev norm~\citep{adams2003sobolev}. Utilizing these, we then give the definition of Sobolev IPM for probability measures on a graph.

\begin{definition}[Graph-based Sobolev space~\citep{le2022st}]\label{def:Sobolev}
 Let $\lambda$ be a nonnegative Borel measure on $\G$, and let  $1\leq p\leq \infty$. A continuous function $f: \G \to \R$ is said to belong to the Sobolev space $W^{1,p}(\G, \lambda)$ if there exists a  function $h\in L^p( \G, \lambda) $ satisfying 
\begin{equation}\label{FTC}
f(x) -f(z_0) =\int_{[z_0,x]} h(y) \lambda(\mathrm{d}y)  \quad \forall x\in \G.
\end{equation}
Such function  $h$ is unique in $L^p(\G, \lambda) $ and is called the graph derivative of $f$ w.r.t.~the measure $\lambda$. The graph derivative of $f \in W^{1,p}(\G, \lambda)$ is denoted as $f' \in L^p( \G, \lambda)$.
\end{definition}

\paragraph{Sobolev norm.} $W^{1,p}(\G, \lambda)$ is a normed space with the Sobolev norm~\citep[\S3.1]{adams2003sobolev} being defined as
\begin{equation}\label{eq:SobolevNorm}
    \norm{f}_{W^{1,p}} := \left( \norm{f}_{L^p}^p + \norm{f'}_{L^p}^p \right)^{\frac{1}{p}}.
\end{equation}
Additionally, let $W^{1, p}_0(\G, \lambda)$ be the subspace consisting of all functions $f$ in $W^{1, p}(\G, \lambda)$ satisfying $f(z_0) = 0$. Denote $\calB(p) := \left\{ f \in W^{1, p}_0(\G, \lambda): \norm{f}_{W^{1, p}} \le 1 \right\}$ as the unit ball in the Sobolev space.

\paragraph{Sobolev IPM.} Sobolev IPM for probability measures on a graph is an instance of the integral probability metric (IPM) where its critic function belongs to the graph-based Sobolev space, and is constrained within the unit ball of that space. More concretely, given a nonnegative Borel measure $\lambda$ on $\G$, an exponent
$1 \le p \le \infty$ and its conjugate $p'$,\footnote{$p' \in [1, \infty]$ satisfying $\frac{1}{p} + \frac{1}{p'} = 1$. If $p=1$, then $p' = \infty$.} the Sobolev IPM between any two probability measures $\mu, \nu \in \calP(\G)$  is defined as
\begin{equation}\label{eq:SobolevIPM}
\calS_{p}(\mu, \nu) := \hspace{-0.4em} \sup_{f \in \calB(p')} \hspace{-0.2em} \left| \int_{\G} f(x) \mu(\text{d}x) - \int_{\G} f(y) \nu(\text{d}y)\right|.
\end{equation}
Notice that the quantity inside the absolute signs is unchanged if $f$ is replaced by $f - f(z_0)$. Thus, we can assume without loss of generality that $f(z_0) = 0$. This is the motivation for our introduction of the Sobolev space $W^{1, p}_0(\G, \lambda)$.\footnote{Similarly, Sobolev space vanishing at the boundary is applied for the Sobolev GAN~\citep{Mroueh-2018-Sobolev}.}

To our knowledge, there are no efficient algorithmic approaches to compute the Sobolev IPM in Equation~\eqref{eq:SobolevIPM} effectively, especially for large-scale settings.

\paragraph{Weight function.} Hereafter, we consider the weight function, defined as
\begin{equation}\label{eq:weighting_func_wLp}
\hat{w}(x) := 1 + \lambda(\Lambda(x)), \qquad \forall x \in \G,
\end{equation}
where recall that $\Lambda(x)$ is a subgraph (see Equation~\eqref{sub-graph}, and Figure~\ref{fg:GeodeticGraph}), and $\lambda(\cdot)$ is measure $\lambda$ on a set. For instance, when $\lambda$ is the length measure on graph $\G$~\citep{le2022st}\footnote{See Appendix \S\ref{app:sec:ST} for a review.}, then $\lambda(\Lambda(x))$ is the total length of $\Lambda(x)$.

The next result plays the key role in our approach.

\begin{theorem}[Equivalence]\label{thrm:S_wLp_norm}
For a nonnegative Borel measure $\lambda$ on $\G$ and  $1 \le p < \infty$, let $c_1 := \left[ \frac{\min(1, \lambda(\G)^{p-1})}{1 + \lambda(\G)^{p}} \right]^{\frac{1}{p}}$, and $c_2 := \left[\max(1,  \lambda(\G)^{p-1})\right]^{\frac{1}{p}}$. Then, we have 
\begin{equation}\label{eq:equivalence_S_wLp}
c_1\norm{f'}_{L^p_{\hat{w}}} \le \norm{f}_{W^{1, p}} \le c_2\norm{f'}_{L^p_{\hat{w}}},
\end{equation}
for every function $f\in W^{1, p}_0(\G, \lambda)$.
\end{theorem}
The proof is placed in Appendix \S\ref{app:sec:proofs:equivalence_SN_wLP}.

Relation \eqref{eq:equivalence_S_wLp} implies that the Sobolev norm of a critic function $f \in W_0^{1, p}(\G, \lambda)$ is equivalent to the weighted $L^p$-norm of its gradient $f'$. Moreover, the weight function is given explicitly by Equation~\eqref{eq:weighting_func_wLp}.


\subsection{Regularized Sobolev IPM}


Based on the equivalent relation given by Theorem~\ref{thrm:S_wLp_norm}, we propose to regularize the Sobolev IPM (Equation~\eqref{eq:SobolevIPM}) by relaxing the constraint on the critic function $f$ in the graph-based Sobolev space $W^{1,p'}_0$. More precisely, instead of $f$ belonging to  the unit ball  $\calB(p')$ of the Sobolev space, we propose to  constraint critic $f$ within the unit ball $\calB(p', \hat{w})$ of the weighted $L^{p'}\!$-norm of $f'$ with weight
function $\hat{w}$. Hereafter, $\calB(p', \hat{w})$ is defined by
\begin{equation}\label{eq:weightedLp_ball_regularized}
\calB(p', \hat{w}) := \left\{f \in W^{1, p'}_0 \! (\G, \lambda) : \norm{f'}_{L^{p'}_{\hat{w}}} \leq 1 \right\}.
\end{equation}
We now formally define the regularized Sobolev IPM between two probability distributions on graph $\G$.
\begin{definition}[Regularized Sobolev IPM on graph]\label{def:regSobolevIPM}
Let $\lambda$ be a nonnegative Borel measure on $\G$ and $1 \le p \le \infty$.  Then for any given probability measures $\mu, \nu \in \calP(\G)$, the regularized Sobolev IPM is defined as
\begin{equation}\label{eq:regSobolevIPM}
\hat{\calS}_{p}(\mu, \nu) \hspace{-0.1em} := \hspace{-1.3em} \sup_{f \in \calB(p', \hat{w})} \hspace{-0.2em} \left| \int_{\G} f(x) \mu(\text{d}x) - \int_{\G} f(y) \nu(\text{d}y)\right|.\hspace{-0.25em}
\end{equation}
\end{definition}




The following result shows that the regularized Sobolev IPM yields a closed-form expression.
\begin{theorem}[Closed-form expression]\label{thrm:closed_form_regSobolevIPM}
Let $\lambda$ be any nonnegative Borel measure on $\G$, and $1 \le p < \infty$.\footnote{See Appendix \S\ref{app:sec:further_results:theory} for $\hat{\calS}_{\infty}$.} Then
\begin{equation}\label{eq:closed_form_regSobolevIPM}
\hat{\calS}_{p}(\mu, \nu)^p \hspace{-0.1em} = \hspace{-0.4em} \int_{\G} \hspace{-0.3em} \hat{w}(x)^{1-p} \left| \mu(\Lambda(x)) - \nu(\Lambda(x)) \right|^p \lambda(\dd x). \hspace{-0.18em}
\end{equation}
\end{theorem}
The proof is placed in Appendix \S\ref{sec:proof_closedform}.

Notice that in identity~\eqref{eq:closed_form_regSobolevIPM}, both the subgraph $\Lambda(x)$ (Equation~\eqref{sub-graph}) and the weight function $\hat{w}(x)$ (Equation~\eqref{eq:weighting_func_wLp}) depend on input point $x$ under the integral over $\G$.

For practical applications, we next derive an explicit  formula for the integral over graph $\G$ in Equation~\eqref{eq:closed_form_regSobolevIPM} 
when the input probability measures are supported on nodes  $V$ of graph $\G$. This gives an  efficient method for computing the regularized Sobolev IPM $\hat{\calS}_p$.
 Note that to achieve this result, we use the length measure on graph $\G$~\citep{le2022st} for the nonnegative Borel measure $\lambda$, i.e., we have $\lambda([x, z]) = d_{\G}(x, z), \forall x, z \in \G$. We summarize the result in the following theorem. 
    
\begin{theorem}[Discrete case]\label{thrm:regSobolevIPM_discrete}
Let $\lambda$ be the length measure on $\G$, and $1 \le p < \infty$. Suppose that $\mu,\nu\in \calP(\G)$ are supported on nodes $V$ of graph $\G$.\footnote{We discuss an extension for measures supported in graph $\G$ in Appendix \S\ref{app:sec:further_discussions}.} 
Then we have
\begin{equation}\label{eq:regSobolevIPM_discrete}
\hat{\calS}_{p}(\mu, \nu) = \left( \sum_{e \in E} \beta_e \left| \mu(\gamma_e) - \nu(\gamma_e) \right|^p
\right)^{\frac{1}{p}},
\end{equation}
where for each edge $e \in E$ of graph $\G$, the scalar number $\beta_e$ is given by
\begin{equation}\label{eq:closed_form_edge_weight_regSobolevIPM}
    \hspace{-0.1em}\beta_e := \left\{
    \begin{array}{ll}
          \log{\left(1 + \frac{w_e}{1 + \lambda(\gamma_e)} \right)} & \mbox{if } p = 2, \hspace{-1.1em}\\
          \frac{\left(1 + \lambda(\gamma_e)+ w_e \right)^{2-p} - \left(1 + \lambda(\gamma_e) \right)^{2-p} }{2-p} & \text{otherwise.}\hspace{-1.1em}
    \end{array} 
    \right. 
    \end{equation}
\end{theorem}
The proof is placed in Appendix \S\ref{app:subsec:proof_thrm:regSobolevIPM_discrete}.

\begin{remark}[Non-physical graph]
We assumed that $\G$ is a physical graph in \S\ref{sec:preliminaries}. Nevertheless, when input measures are supported on nodes $V$ of graph $\G$, and $\lambda$ is a length measure of graph $\G$, the regularized Sobolev IPM $\hat{\calS}_{p}$ does not depend on this physical assumption as illustrated in Theorem~\ref{thrm:regSobolevIPM_discrete}. Indeed, it only depends on the graph structure $(V, E)$ and edge weights $w_e$. Therefore, we can apply the regularized Sobolev IPM for non-physical graph $\G$.
\end{remark}

\paragraph{Preprocessing.} For the computation of the regularized Sobolev IPM $\hat{\calS}_{p}$ in Equation~\eqref{eq:regSobolevIPM_discrete}, observe that the set $\gamma_e$ (see Equation~\eqref{sub-graph}) and $\beta_e$ (see Equation~\eqref{eq:closed_form_edge_weight_regSobolevIPM}) can be precomputed for all edges $e$ in $\G$. Notably, the preprocessing procedure is only involved graph $\G$,  and it is independent of the input probability measures. Consequently, we only need to perform this preprocessing procedure once, regardless of the number of input pairs of probability measures that are evaluated by the regularized Sobolev IPM $\hat{\calS}_{p}$ in applications. More precisely, we apply the Dijkstra algorithm to recompute the shortest paths from root node $z_0$ to all other input supports (or vertices) with complexity $\mathcal{O}(|E| + |V| \log{|V|})$ where we write $|\cdot|$ for the set cardinality. Then, we can evaluate $\gamma_e$ and $\beta_e$ for each edge $e$ in $E$.

\paragraph{Sparsity of subgraph $\gamma_e$ in $\G$.} Denote $\text{supp}(\mu)$ as the set of supports of measure $\mu$. For any $x \in \text{supp}(\mu)$, its mass is accumulated into $\mu(\gamma_e)$ if and only if $e \subseteq [z_0, x]$~\citep{le2022st}. Therefore, let define $E_{\mu, \nu} \subseteq E$ as follows
\[
E_{\mu, \nu} \hspace{-0.2em}:=\hspace{-0.2em} \left\{e \! \in \! E \mid \exists z \! \in \! (\text{supp}(\mu) \cup \text{supp}(\nu)), e \subseteq [z_0, z] \right\}.
\]
Then, in fact, we can remove all edges $e \in E \setminus E_{\mu, \nu}$ in the summation in Equation~\eqref{eq:regSobolevIPM_discrete} for the computation of the regularized Sobolev IPM $\hat{\calS}_{p}$. As the result of this sparsity property, the computational complexity of $\hat{\calS}_{p}$ is reduced to $\mathcal{O}(|E_{\mu, \nu}|)$.

Following the preprocessing procedure and the sparsity of subgraph $\gamma_e$ for any edge $e$ in $\G$, we only need to check the precomputed shortest paths from root node $z_0$ to each support of input measures $\mu, \nu$ to respectively obtain $\mu(\gamma_e), \nu(\gamma_e)$ for all edges $e \in E_{\mu, \nu}$ in graph $\G$. Additionally, $\beta_e$ is also precomputed for all edge $e$ in $\G$. Therefore, it is straightforward to implement the computation of $\hat{\calS}_{p}(\mu, \nu)$ in Equation~\eqref{eq:regSobolevIPM_discrete} with complexity $\mathcal{O}(|E_{\mu, \nu}|)$.

\section{Properties of Regularized Sobolev IPM}
\label{sec:properties_regSobolevIPM}

\subsection{Metric and Its Relations}\label{subsec:metric_relations}

\paragraph{Metrization.} We prove that the regularized Sobolev IPM is a metric.

\begin{theorem}[Metrization] \label{thrm:metrize}
For any $1 \le p \le \infty$, the regularized Sobolev IPM $\hat{\calS}_p$ is a metric on the space $\calP(\G)$
of probability measures on graph $\G$.
\end{theorem}

The proof is placed in Appendix \S\ref{app:subsec:proof_thrm:metrize}.


\paragraph{Connection to the original Sobolev IPM.} We next show that the regularized Sobolev IPM is equivalent to the original Sobolev IPM.
\begin{theorem}[Relation with original Sobolev IPM]\label{thrm:relationSobolevIPM}
For a nonnegative Borel measure $\lambda$ on $\G$, $1 \le p < \infty$, 
we have
\begin{equation}\label{eq:equivalence_SobolevIPM}
c_1 \, \calS_p(\mu, \nu) \le \hat{\calS}_p(\mu, \nu) \le c_2 \, \calS_p(\mu, \nu)
\end{equation}
for every $\mu, \nu \in \calP(\G)$, where $c_1, c_2$ are constants defined in Theorem~\ref{thrm:S_wLp_norm}. Hence, the regularized Sobolev IPM is equivalent to the original Sobolev IPM.
\end{theorem}

The proof is placed in Appendix \S\ref{app:subsec:proof_thrm:relationSobolevIPM}.



\paragraph{Connection to the Sobolev transport~\citep{le2022st}.} 
For $1 \le p < \infty$, let $\mathcal{ST}_{\!p}$ be the $p$-order Sobolev transport, which is a scalable variant of optimal transport for measures on a graph~\citep{le2022st}.\footnote{We review Sobolev transport $\mathcal{ST}_{\!p}$ in Appendix \S\ref{app:subsec:review-ST}.} The next result provides the relation between our regularized Sobolev IPM $\hat\calS_p$ and the Sobolev transport $\mathcal{ST}_{\!p}$.

\begin{proposition}[Relation with Sobolev transport]\label{prop:relationSobolevTransport}
For a nonnegative Borel measure $\lambda$ on $\G$, $1 \le p < \infty$, we have
\begin{equation}\label{eq:equivalence_SobolevTransport}
(1 + \lambda({\G}))^{\frac{1-p}{p}} \mathcal{ST}_{\!p}(\mu, \nu) \le  \hat{\calS}_p(\mu, \nu) \le \mathcal{ST}_{\!p}(\mu, \nu)
\end{equation}
for every  $\mu, \nu \in \calP(\G)$. Hence, the regularized Sobolev IPM is equivalent to the Sobolev transport.
\end{proposition}

The proof is placed in Appendix \S\ref{app:subsec:proof_ST_relation}.


\paragraph{Regularized Sobolev IPM with different orders.} We establish a relation between different orders for the regularized Sobolev IPM.

\begin{proposition}[Upper bound] \label{prop:upper}
Let $\lambda$ be any finite and nonnegative Borel measure  on $\G$. Then we have
\[
\hat{\calS}_p(\mu, \nu) \le
\left[ \lambda(\G) \left(1 +\lambda(\G) \right) \right]^{\frac{1}{p} - \frac{1}{q}}   \hat{\calS}_{q}(\mu, \nu)
\]
for any exponents $p$ and $q$ satisfying $1\leq p < q < \infty$.
\end{proposition}

The proof is placed in Appendix \S\ref{app:subsec:proof-different-pq}.

\paragraph{Connection to the Wasserstein distance.} We establish the following relation between the regularized Sobolev IPM and the Wasserstein distance.

\begin{proposition}[Tree topology] \label{prop:tree}
Suppose that graph $\G$ is a tree and $\lambda$ is the length measure on graph $\G$. Then, for probability measures $\mu, \nu \in \calP(\G)$, we have 
\[
\hat{\calS}_1(\mu, \nu) = \calW_1(\mu, \nu),
\]
where $\calW_1$ is the $1$-Wasserstein distance\footnote{We review $p$-Wasserstein distance $\calW_p$ in Appendix \S\ref{app:subsec:review-IPM-OT}.} with ground cost $d_{\G}$ (i.e., tree-Wasserstein distance~\citep{LYFC}).
\end{proposition}

The proof is placed in Appendix \S\ref{app:subsec:proof-relation-regSIPM1-W1}.

We note that it is still an open question for the exact relationship between the regularized Sobolev IPM $\hat{\calS}_p$ and the $p$-Wasserstein distance $\calW_p$ when $p>1$. Nevertheless, we next illustrate that $\hat{\calS}_p$ is lower bounded by $\calW_1$.

\begin{proposition}[Bounds]\label{prop:w1-vs-sp}
Suppose that graph $\G$ is a tree and $\lambda$ is the length measure on graph $\G$. Then, for any $1 \le p < \infty$, we have
\[
\hat\calS_p(\mu,\nu ) \geq \left[ \lambda(\G)(1 + \lambda(\G)) \right]^{\frac{1-p}{p}} \calW_1(\mu, \nu).
\]
\end{proposition}

The proof is placed in Appendix \S\ref{app:subsec:proof-relation-regSIPMp-W1}.


\subsection{Regularized Sobolev IPM Kernels}\label{subsec:kernels_SIPM}

\paragraph{Negative definiteness.\footnote{We follow the definition of negative definiteness in \citet[pp. 66--67]{Berg84}. We give a review on kernels in Appendix \S\ref{app:subsec:review-kernels}.}}


\begin{proposition}[Negative definiteness]\label{prop:neg_def}
 Suppose that $\lambda$ is the length measure on graph $\G$, and input probability measures are supported on nodes $V$ of graph $\G$. Then $\hat\calS_p$ and $\hat\calS_p^p$ are negative definite for every $1 \le p \le 2$.
\end{proposition}

The proof is placed in Appendix \S\ref{app:subsec:proof-negdef}.

\paragraph{Positive definite kernels.}

From Proposition~\ref{prop:neg_def} and following 
\citet[Theorem~3.2.2, pp.~74]{Berg84}, given $t > 0$, $1 \le p \le 2$ and probability measures $\mu, \nu$ supported on nodes $V$ of graph $\G$, the kernels 
\begin{eqnarray}
&k_{\hat\calS_p}(\mu, \nu) \coloneqq \exp(-t \hat\calS_p(\mu, \nu)), \\
&k_{\hat\calS^p_p}(\mu, \nu) \coloneqq \exp(-t \hat\calS_p(\mu, \nu)^p)
\end{eqnarray}
are positive definite. 

\paragraph{Infinite divisibility for the regularized Sobolev IPM kernels.} We show that the kernels $k_{\hat\calS_p}$ and $k_{\hat\calS^p_p}$ based on the regularized Sobolev IPM are infinitely divisible.
\begin{proposition}[Infinite divisibility]\label{prop:divisibility}
For $1 \le p\le 2$, the kernels $k_{\hat\calS_p}$ and $k_{\hat\calS^p_p}$ are infinitely divisible.
\end{proposition}

The proof is placed in Appendix \S A.12.

As for infinitely divisible kernels, regardless of their hyperparameter $t$, it suffices to compute the Gram matrix of these kernels $k_{\hat\calS_p}$ and $k_{\hat\calS^p_p}$ ($1 \le p\le 2$) for probability measures on a graph in the training set once.


\section{Related Works}
\label{sec:related_works}

In this section, we discuss related works capitalizing  on the Sobolev IPM and Sobolev geometric structure.

\citet{liang2017well_arXiv} leverages Sobolev IPM to study the convergence rate for learning density with the GAN framework. This work shows how convergence rates depend on Sobolev smoothness restrictions within the Sobolev IPM.~\citet{liang2017well_arXiv} also further exploits Sobolev geometry to derive generalization bounds for deep ReLU discriminator networks in GAN. Additionally,~\citet{singh2018nonparametric, liang2021well_jmlr} improve these results under Sobolev IPM by employing the adversarial framework for the analysis. More recently,~\citet{kozdobasobolev} leverages Sobolev norm for unnormalized density estimation.



\citet{nickl2007bracketing} study bracket metric entropy for Sobolev space, which plays a central role in many limit theorems for empirical processes~\citep{dudley1978central, ossiander1987central, andersen1988central}, and for studying convergence rates, lower risk bounds of statistical estimators~\citep{birge1993rates, geer2000empirical}.

Sobolev GAN~\citep{Mroueh-2018-Sobolev} exploits the so-called Sobolev discrepancy to compare probability measures for the GAN framework. More concretely, Sobolev GAN constraints the critic function in IPM within a unit ball defined by the $L^2$-norm of its gradient function with respect to a dominant measure, which shares the same spirit as the Sobolev Wasserstein GAN approach~\citep{xu2020towards}. On the other hand, Fisher GAN~\citep{mroueh2017fisher} constraints the critic function by the $L^2$-norm of itself with respect to a dominant measure. Thus, the Sobolev norm (Equation~\eqref{eq:SobolevNorm}), which integrates information from both critic function and its gradient function, can be regarded as a unification for the approaches in Fisher GAN and Sobolev GAN within the Sobolev IPM problem.

\citet{mroueh2018regularized} proposes the kernelized approach for the Sobolev discrepancy which constraints the critic function of the Sobolev discrepancy within a reproducing kernel Hilbert space. Then \citet{mroueh2019sobolev} leverages the kernel Sobolev discrepancy to quantify kinetic energy to propose Sobolev descent, i.e., a gradient flow that finds a path of distributions from source to target measures minimizing the kinetic energy. Additionally,~\citet{mroueh2020unbalanced} extend the kernelized approach to unbalanced settings where source and target measures may have different total mass.

\citet{belkin2006manifold} leverages Sobolev norm for manifold regularization in semi-supervised learning (SSL).~\citet{husain2020distributional} studies Sobolev IPM uncertainty set for the distributional robust optimization, and links the distributional robustness with the manifold regularization penalty~\citep{belkin2006manifold}. Additionally,~\citet{mroueh2019sobolev_IC} uses Sobolev discrepancy to propose Sobolev independence criterion for nonlinear feature selection. Furthermore,~\citet{nietert2021smooth} employs Sobolev IPM to analyze theoretical properties for Gaussian-smoothed $p$-Wasserstein distance.


Sobolev transport (ST)~\citep{le2022st} is a scalable variant of OT on a graph, which constraints the Lipschitz condition within the graph-based Sobolev space. Moreover, the $p$-order ST can be considered as a generalization of the Sobolev discrepancy, which constraints the critic function within a unit norm defined by the $L^p$-norm of its gradient. Additionally,~\citet{le2023scalable} extend ST for the unbalanced setting where input measures may have different total mass, while ~\citet{le2024generalized} leverage a class of convex functions to extend ST to more general geometric structures which is beyond its original $L^p$-geometric structure.

\begin{figure*}[h]
  \begin{center}
\includegraphics[width=0.82\textwidth]{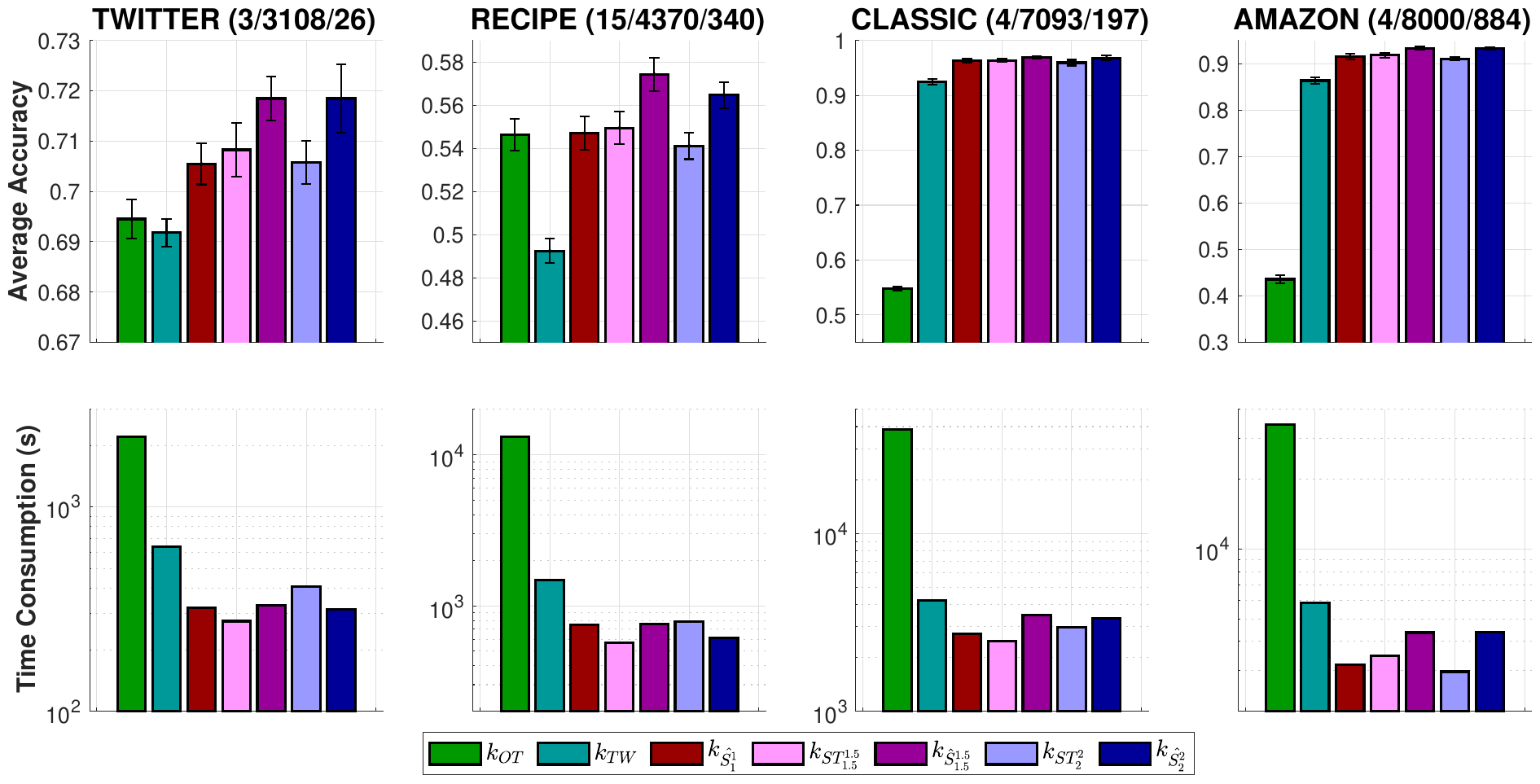}
  \end{center}
  \vspace{-12pt}
  \caption{SVM results and time consumption for kernel matrices with graph $\G_{\text{Log}}$. For each dataset, the numbers in the parenthesis are the number of classes; the number of documents; and the maximum number of unique words for each document respectively.}
  \label{fg:DOC_10KLog_main}
\end{figure*}

\begin{figure*}[h]
  \begin{center}
    \includegraphics[width=0.82\textwidth]{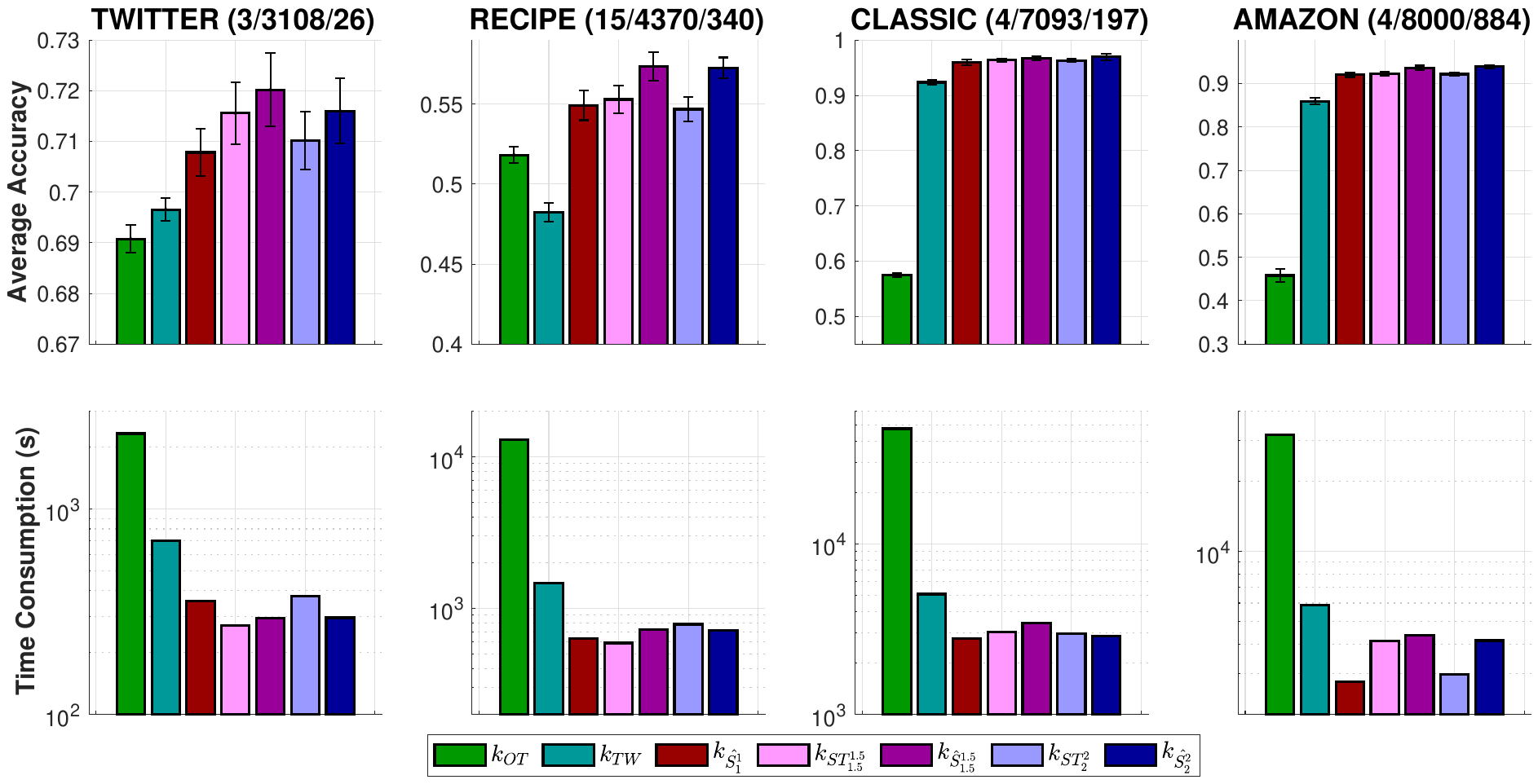}
  \end{center}
  \vspace{-12pt}
  \caption{SVM results and time consumption for kernel matrices with graph $\G_{\text{Sqrt}}$.}
  \label{fg:DOC_10KSqrt_main}
\end{figure*}

\section{Experiments}
\label{sec:experiments}

In this section, we illustrate the fast computation for the regularized Sobolev IPM, which is comparable to the Sobolev transport (ST), and several-order faster than the standard optimal transport (OT) for measures on a graph. We then show preliminary evidences on the advantages of the regularized Sobolev IPM kernels to \emph{compare probability measures on a given graph under the same settings} for document classification and for TDA. 

\paragraph{Document classification.} We consider $4$ popular document datasets: \texttt{TWITTER, RECIPE, CLASSIC, AMAZON}. The properties of these datasets are summarized in Figure~\ref{fg:DOC_10KLog_main}. As in~\citet{le2022st}, we use word2vec word embedding~\citep{mikolov2013distributed} to map words into vectors in~$\R^{300}$. Thus, each document is represented as a probability measure where its supports are in $\R^{300}$, and their corresponding mass is the word frequency in the document.

\begin{figure}[h]
  \begin{center}
    \includegraphics[width=0.48\textwidth]{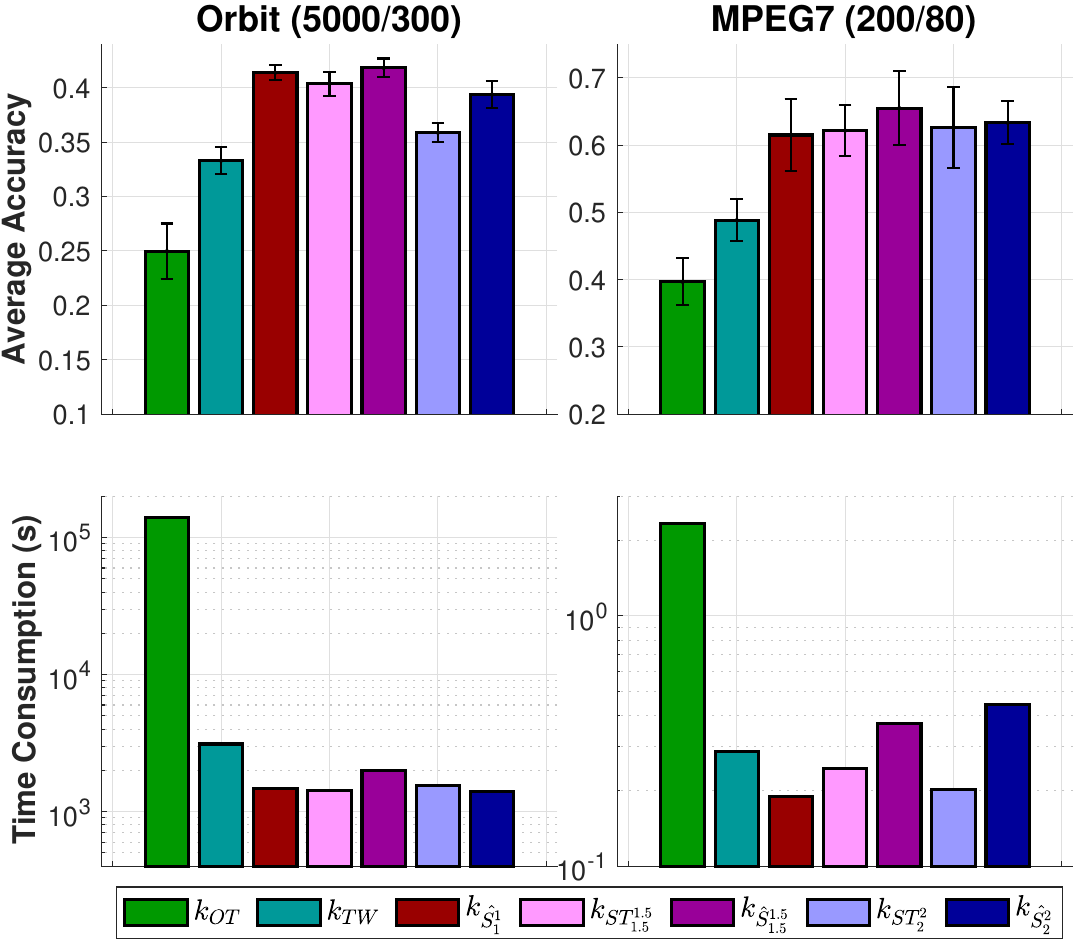}
  \end{center}
  \vspace{-6pt}
  \caption{SVM results and time consumption for kernel matrices with graph $\G_{\text{Log}}$. For each dataset, the numbers in the parenthesis are respectively the number of PD; and the maximum number of points in PD.}
  \label{fg:TDA_mix10KLog_main}
\end{figure}

\paragraph{TDA.} We consider orbit recognition on the synthesized \texttt{Orbit} dataset~\citep{adams2017persistence}, and object classification on a $10$-class subset of \texttt{MPEG7} dataset~\citep{latecki2000shape} as in \citet{le2022st}. The properties of these datasets are shown in Figure~\ref{fg:TDA_mix10KLog_main}. Objects of interest are represented by persistence diagrams (PD), extracted by algebraic topology methods (e.g., persistence homology)~\citep{edelsbrunner2008persistent}. Therefore, each PD can be regarded as a distribution of 2D topological feature data points with a uniform mass.

\paragraph{Graph.} We consider the graphs $\G_{\text{Log}}$ and $\G_{\text{Sqrt}}$~\citep[\S5]{le2022st} for experiments. Empirically, these graphs satisfy the assumptions in \S\ref{sec:preliminaries}, similar to observations in~\citet{le2022st}.\footnote{We review these graphs in Appendix~\S\ref{app:sec:further_discussions}.} We consider $M\!=\! 10^2, 10^3, 10^4$ as the number of nodes for these graphs, but omit $M=10^4$ in \texttt{MPEG7} due to the small size of this dataset. Recall that graphs $\G_{\text{Log}}$ and $\G_{\text{Sqrt}}$ have $(M\log{M})$ and $M^{3/2}$ edges respectively.

\begin{table}[ht]
\caption{The number of pairs of probability measures on datasets for kernelized SVM.}
\label{tb:numpairs}
    \centering
    \vspace{-6pt}
\begin{tabular}{|l|c|}
\hline
Datasets & \#pairs \\ \hline
\texttt{TWITTER}  & 4394432 \\ \hline
\texttt{RECIPE}   & 8687560      \\ \hline
\texttt{CLASSIC}  & 22890777       \\ \hline
\texttt{AMAZON}   & 29117200      \\ \hline
\texttt{Orbit}    & 11373250   \\ \hline
\texttt{MPEG7}    & 18130     \\ \hline
\end{tabular}
\vspace{-13pt}
\end{table}

\paragraph{Root node $z_0$.} The regularized Sobolev IPM $\hat{\calS}$ is defined on graph $\G$ with root node $z_0$. Notice that $z_0$ characterizes the graph derivative for functions on $\G$ (Definition~\ref{def:Sobolev}). Much as Sobolev transport~\citep{le2022st}, we employ the sliced approach to weaken this dependency. More precisely, we uniformly average $\hat{\calS}$ over different choices of root nodes, i.e., a sliced variant for~$\hat{\calS}$.

\begin{figure*}[ht]
  \begin{center}
\includegraphics[width=0.8\textwidth]{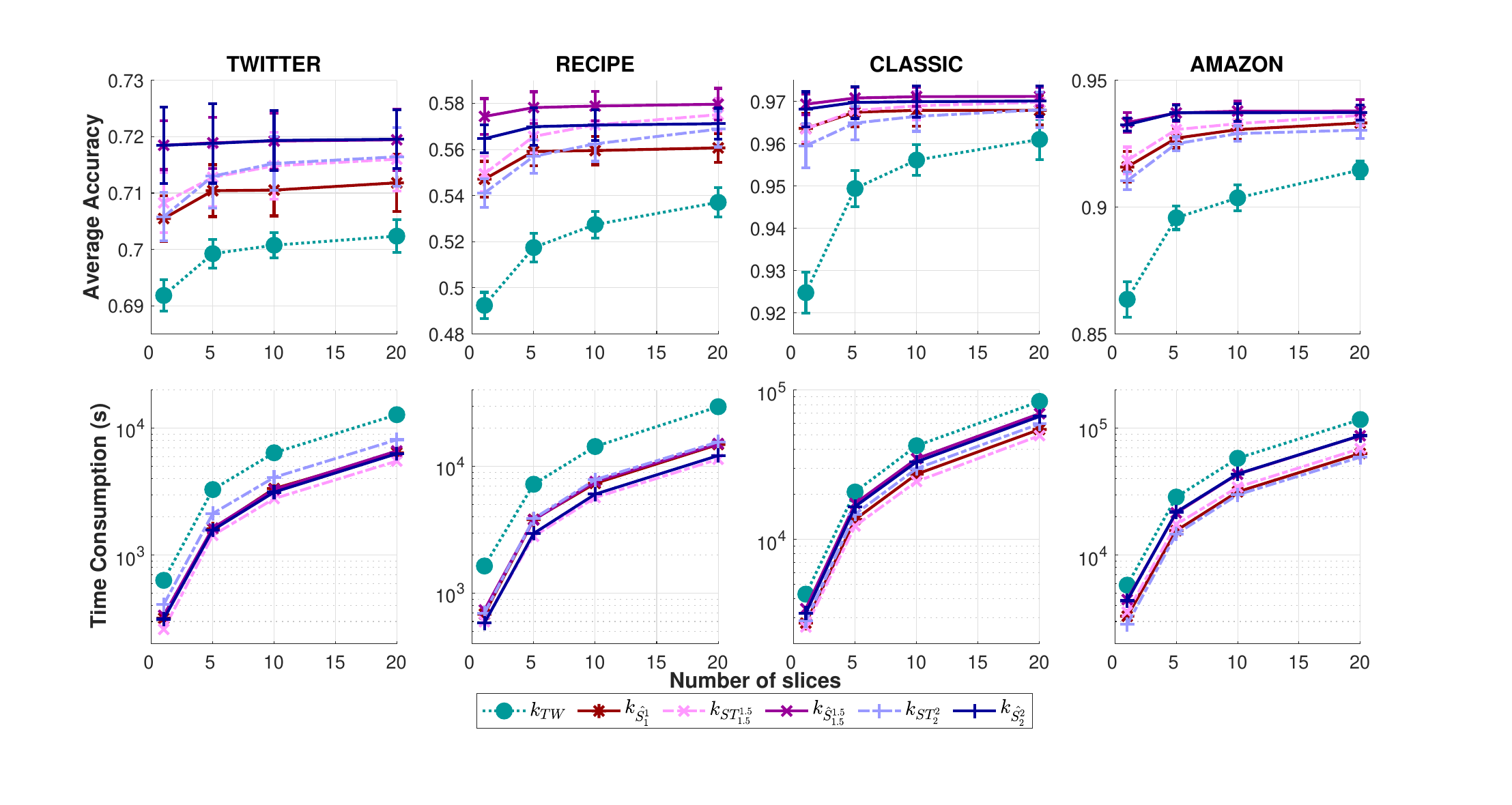}
  \end{center}
  \vspace{-12pt}
  \caption{SVM results and time consumption for kernel matrices of slice variants with graph $\G_{\text{Log}}$.}
  \label{fg:DOC_10KLog_SLICE_main}
 \vspace{-8pt}
\end{figure*}

\paragraph{Classification.} We use kernelized support vector machine (SVM) for document classification and TDA. Specifically, we evaluate regularized Sobolev IPM kernels $k_{\hat\calS_1}, k_{\hat\calS_{1.5}^{1.5}}, k_{\hat\calS_2^2}$; ST kernels $k_{\mathcal{ST}_{1.5}^{1.5}}, k_{\mathcal{ST}_2^2}$,\footnote{Note that $k_{\mathcal{ST}_{\!1}} = k_{\hat\calS_1}$ by Proposition~\ref{prop:relationSobolevTransport}.}; and kernels $\exp(-t \bar{d}(\cdot, \cdot))$ with $t > 0$, where $\bar{d}$ is a distance (e.g., OT with ground cost $d_{\G}$, tree-Wasserstein (TW) with tree sampled from graph $\G$) as considered in~\citet{le2022st}. The kernels corresponding to the two  choices of $\bar{d}$ are respectively denoted as $k_{OT}, k_{TW}$. Note that kernel $k_{OT}$ is empirically indefinite. We regularize its Gram matrices by adding a sufficiently large diagonal term as in~\citet{Cuturi-2013-Sinkhorn}.

We randomly split each dataset into $70\%/30\%$ for training and test respectively, with $10$ repeats, and use 1-vs-1 strategy for SVM classification. Typically, hyper-parameters are chosen via cross validation. Concretely, SVM regularization is chosen from $\left\{0.01, 0.1, 1, 10\right\}$, and kernel hyperparameter is chosen from $\{1/q_{s}, 1/(2q_{s}), 1/(5q_{s})\}$ with $s = 10, 20, \dotsc, 90$, where we write $q_s$ for the $s\%$ quantile of a subset of corresponding distances on training set. The reported time consumption for kernel matrices includes all preprocessing procedures, e.g., computing shortest paths on graph for $\hat{\calS}$, $\mathcal{ST}$, or sampling tree from graph for TW.

In Table~\ref{tb:numpairs}, we show the number of pairs requiring to evaluate distance for SVM on each run. Especially, in \texttt{AMAZON}, there are more than $29$ million pairs of probability measures.

\begin{figure}[h]
  \begin{center}
    \includegraphics[width=0.48\textwidth]{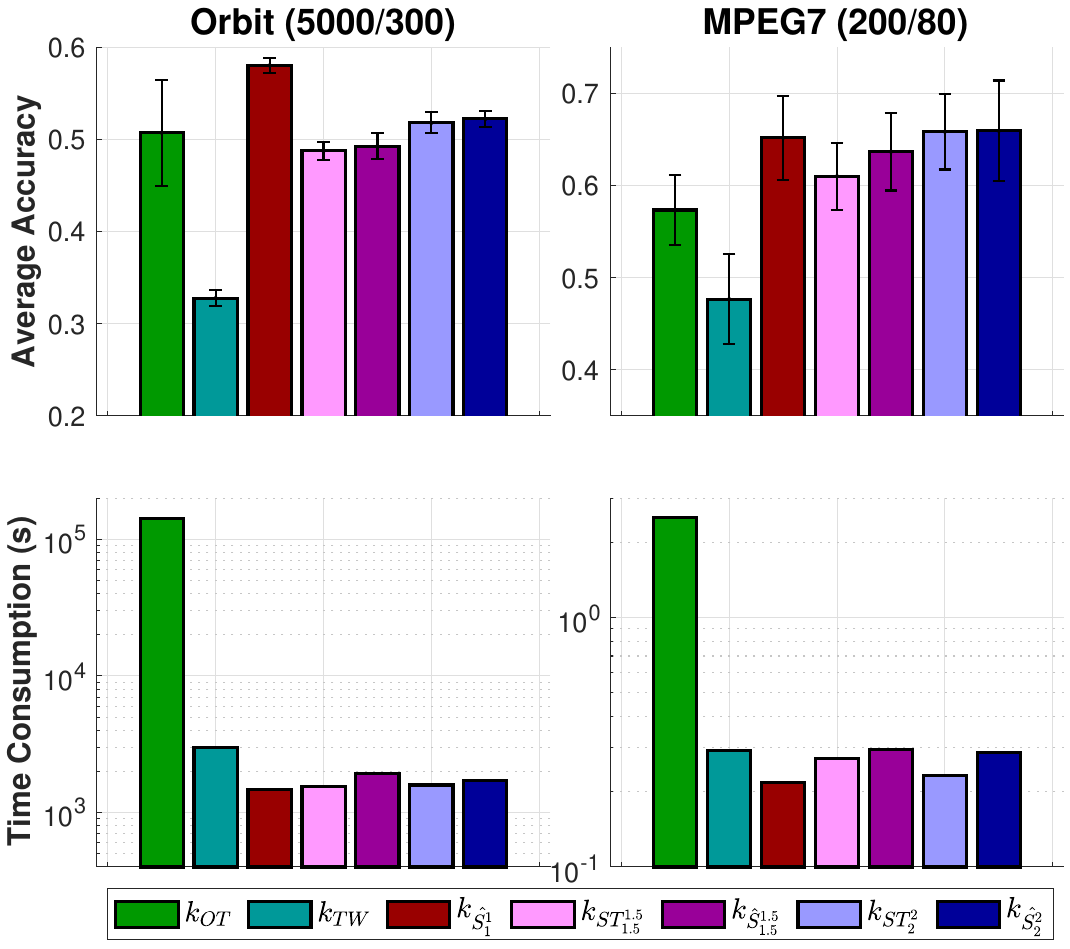}
  \end{center}
  \vspace{-6pt}
  \caption{SVM results and time consumption for kernel matrices with graph $\G_{\text{Sqrt}}$.}
  \label{fg:TDA_mix10KSqrt_main}
\end{figure}

\paragraph{SVM results and discussions.} The reported results for all datasets are with $M = 10^4$, except for \texttt{MPEG7} with $M=10^3$ due to the small size of this dataset. We illustrate SVM results and time consumption for kernel matrices on document classification for graphs $\G_{\text{Log}}$ and $\G_{\text{Sqrt}}$ in Figures~\ref{fg:DOC_10KLog_main} and~\ref{fg:DOC_10KSqrt_main}  respectively. For TDA, we show the results for $\G_{\text{Log}}$ and $\G_{\text{Sqrt}}$ in Figures~\ref{fg:TDA_mix10KLog_main} and~\ref{fg:TDA_mix10KSqrt_main}  respectively.

The computation of regularized Sobolev IPM $\hat{\calS}$ is several-order faster than the traditional OT with ground cost $d_{\G}$. Additionally, the computation of $\hat{\calS}$ is comparable to Sobolev transport $\mathcal{ST}$, a.k.a., a scalable variant of OT on a graph. Especially, for \texttt{Orbit} dataset, either with graph $\G_{\text{Log}}$ or $\G_{\text{Sqrt}}$, it took more than \emph{$39$ hours} to evaluate the kernel Gram matrices for the OT kernel $k_{OT}$, but less than \emph{$34$ minutes} for regularized Sobolev IPM kernels for all exponents $p = 1, 1.5, 2$.

The performances of regularized Sobolev IPM kernels compare favorably with those of ST, OT, TW kernels. Similar to observations in~\citet{le2022st}, the infiniteness of $k_{OT}$ may become a hinder for its performances in certain applications. For examples, the performances of $k_{OT}$ are affected in most of our experiments, except the ones in \texttt{RECIPE} with graph $\G_{\text{Log}}$ and in \texttt{Orbit} with graph $\G_{\text{Sqrt}}$. Additionally, TW kernel $k_{TW}$ uses a partial graph information, but is positive definite unlike its counterpart $k_{OT}$. Performances of $k_{TW}$ are worse than its counterpart $k_{OT}$ in \texttt{RECIPE}, but are better than $k_{OT}$ in \texttt{CLASSIC}, \texttt{AMAZON} for both graphs $\G_{\text{Log}}$ and $\G_{\text{Sqrt}}$, which agrees with observations in~\citet{le2022st}. For performances, similar to Sobolev transport, one may turn the exponent $p$ for the regularized Sobolev IPM in applications, e.g., via cross validation. 

\begin{figure}[H]
  \begin{center}
    \includegraphics[width=0.48\textwidth]{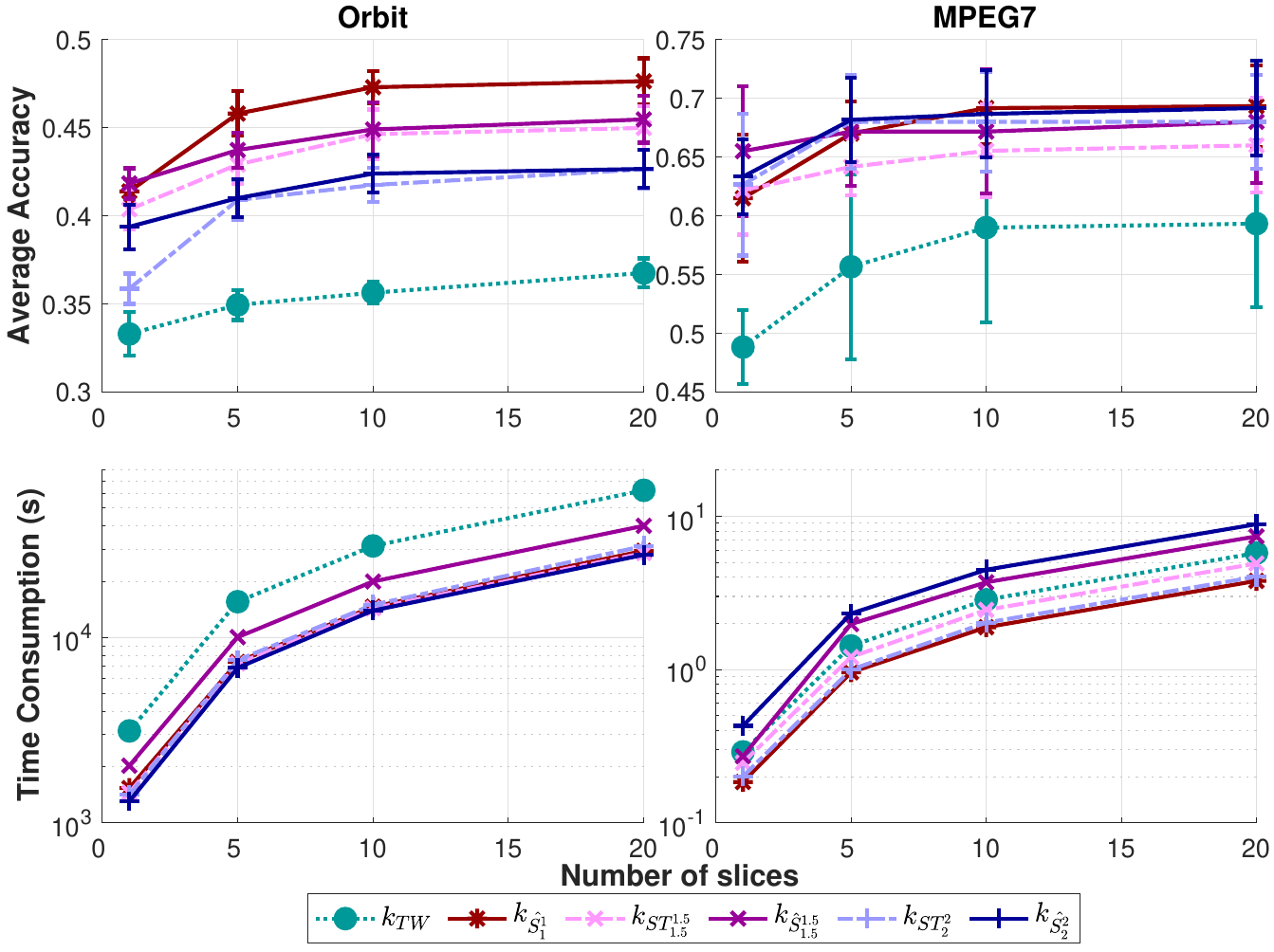}
  \end{center}
  \vspace{-6pt}
  \caption{SVM results and time consumption for kernel matrices of slice variants with $\G_{\text{Log}}$.}
  \label{fg:TDA_mix10K1K_Log_AccTime_SLICE}
\end{figure}

Additionally, we illustrate performance of sliced variants with graph $\G_{\text{Log}}$ in Figures~\ref{fg:DOC_10KLog_SLICE_main} and~\ref{fg:TDA_mix10K1K_Log_AccTime_SLICE} for document classification and TDA respectively. When the number of slices increases, their performances are improved but it comes with a trade-off on increasing the computational time linearly.

Further empirical results are placed in Appendix \S\ref{app:sec:further_results:exp}.


\section{Conclusion}
\label{sec:conclusion}

In this work, we propose a novel regularization for Sobolev IPM for probability measures on a graph. The regularized Sobolev IPM admits a closed-form expression for fast computation. It paves the way for applying Sobolev IPM in applications, especially for large-scale settings. For future works, it is interesting to extend the approach to unbalanced setting where input measures may have different total mass, and to go beyond the Sobolev geometric structure, e.g. tackling more challenging geometric structure for IPM such as critic function within a unit ball defined by Besov norm, i.e., Besov IPM, for practical applications.

\section*{Acknowledgements}

We thank the area chairs and anonymous reviewers for their comments. KF has been supported in part by Grant-in-Aid for Transformative Research Areas (A) 22H05106 and JST CREST JPMJCR2015. HH acknowledges the support of JSPS KAKENHI JP25H01494 and JP23K24909. TL gratefully acknowledges the support of JSPS KAKENHI Grant number 23K11243, and Mitsui Knowledge Industry Co., Ltd. grant.

%


\section*{Impact Statement}

The paper proposes a novel regularization for Sobolev IPM for probability measures on a graph, which yields a closed form expression for fast computation. Our work paves a way to use Sobolev IPM for practical applications, especially for large-scale settings. To our knowledge, there are no foresee potential societal consequences of our work.




\balance


\bibliographystyle{icml2024}
\bibliography{bibEPT21, bibSobolev22}

\newpage
\appendix
\onecolumn

\begin{center}
{\bf{\Large{\textit{Supplement to}  ``Scalable Sobolev IPM for Probability Measures on a Graph"}}}
\end{center}

The supplementary is organized into three parts as follows:
\begin{itemize}
\item In Section~\ref{app:sec:proofs}, we provide the proofs for the theoretical results in the main manuscript.

\item In Section~\ref{app:sec:reviews}, we briefly review related notions used in our work.

\item In Section~\ref{app:sec:further_results}, we provide further experimental results and discussions about our proposed approach.
\end{itemize}

\paragraph{Notations.} Let $\mathbf{1}_{X}(x)$ be the indicator function, i.e., 
\[ 
\mathbf{1}_{X}(x)= \left\{
\begin{array}{ll}
      1 & \text{if }x \in X \\
      0 & \text{otherwise.}
\end{array} 
\right. 
\]
For two points $u, v \in \R^n$, let $\langle u,  v\rangle$ be the line segment in $\R^n$ connecting the two points $u, v$, and denote $( u, v)$ for the same line segment but without its two end-points.

\section{Proofs}\label{app:sec:proofs}

In this section, we give the detailed proofs for the theoretical results in the main manuscript.

\subsection{Proof for Theorem~\ref{thrm:S_wLp_norm}}\label{app:sec:proofs:equivalence_SN_wLP}

\begin{proof}

Let $f \in W_0^{1, p}$.
We first derive an upper bound for $\norm{f}_{L^p}^p$ in terms of $\norm{f'}_{L^p_{\hat{w}_p}}$. Since $f(z_0) = 0$, we have
\begin{eqnarray*}
\norm{f}_{L^p}^{p} &&= \int_{\G} \left|f(x) \right|^p \lambda (\dd x) \nonumber \\
&&= \int_{\G} \left| f(z_0) + \int_{[z_0, x]} f'(y) \lambda(\dd y) \right|^p \lambda(\dd x) \nonumber\\
&&= \int_{\G} \left|\int_{\G} \mathbf{1}_{[z_0, x]}(y)f'(y)\lambda(\dd y) \right|^p \lambda(\dd x) \nonumber\\
&&= \lambda(\G)^p  \int_{\G} \left| \frac{1}{\lambda(\G)}\int_{\G} \mathbf{1}_{[z_0, x]}(y)f'(y)\lambda(\dd y) \right|^p \lambda(\dd x). \nonumber 
\end{eqnarray*}

Therefore, by applying Jensen's inequality, we obtain
\begin{eqnarray*}
\norm{f}_{L^p}^{p} &&\leq \lambda(\G)^p \int_{\G}  \left( \frac{1}{\lambda(\G)}\int_{\G} \Big| \mathbf{1}_{[z_0, x]}(y)f'(y)\Big|^p \lambda(\dd y) \right) \lambda(\dd x) \\
&& = \lambda(\G)^{p-1}  \int_{\G}\int_{\G}  \Big| \mathbf{1}_{[z_0, x]}(y)f'(y)\Big|^p \lambda(\dd y) \lambda(\dd x).
\end{eqnarray*}

By Fubini's theorem, we can interchange the order of the integration. As a consequence, we obtain
\begin{eqnarray}\label{app:eq:JensenInequality_fpp}
\norm{f}_{L^p}^{p} \hspace{-1.7em} 
&&\leq  \lambda(\G)^{p-1}\int_{\G} \int_{\G} \left|\mathbf{1}_{[z_0, x]}(y)f'(y)\right|^p  \lambda(\dd x) \lambda(\dd y) \nonumber \\
&&=  \lambda(\G)^{p-1}\int_{\G} \left(\int_{\G} \mathbf{1}_{[z_0, x]}(y)  \lambda(\dd x)\right) |f'(y)|^p\lambda(\dd y) \nonumber \\
&&= \lambda(\G)^{p-1} \int_{\G} \left|f'(y) \right|^p  \lambda(\Lambda(y))  \lambda(\dd y),
\end{eqnarray}
where we recall that $\Lambda(y) := \left\{x \in \G : y \in [z_0, x] \right\}$ (see Equation~\eqref{sub-graph}).
Due to estimate \eqref{app:eq:JensenInequality_fpp}, we have
\begin{eqnarray}
    \norm{f}_{W^{1, p}} &&= \left( \norm{f}_{L^p}^p + \norm{f'}_{L^p}^p \right)^{\frac{1}{p}} \nonumber \\
    && \leq \left( \lambda(\G)^{p-1} \int_{\G} \left|f'(x) \right|^p  \lambda(\Lambda(x)) \lambda(\dd x) + \int_{\G} |f'(x)|^p \lambda(\dd x) \right)^{\frac{1}{p}} \nonumber \\
    &&= \left( \int_{\G} \Big[ 1 + \lambda(\G)^{p-1}\lambda(\Lambda(x)) \Big] |f'(x)|^p \lambda(\dd x) \right)^{\frac{1}{p}} \nonumber \\
    &&\le \left( \int_{\G} \max(1, \lambda(\G)^{p-1}) \Big[ 1 + \lambda(\Lambda(x)) \Big] |f'(x)|^p \lambda(\dd x) \right)^{\frac{1}{p}} \nonumber \\ 
    &&= c_2 \norm{f'}_{L^p_{\hat{w}}}, \label{app:eq:upper_bound_S_wLp}
\end{eqnarray}
where we recall that $\hat{w}(x) := 1 + \lambda(\Lambda(x)), \forall x \in \G$ (see Equation~\eqref{eq:weighting_func_wLp}), and $c_2 = \left[\max(1,  \lambda(\G)^{p-1})\right]^{\frac{1}{p}}$.

We next derive a lower bound for $\norm{f}_{L^p}^p$ in terms of $\norm{f'}_{L^p_{\hat{w}_p}}$  as follows
\begin{eqnarray}
    \norm{f}_{W^{1, p}} &&= \left( \norm{f}_{L^p}^p + \norm{f'}_{L^p}^p \right)^{\frac{1}{p}} \nonumber \\
    && \ge \norm{f'}_{L^p} \label{app:eq:nonnegativeLp} \\
    && = \left(\int_{\G} |f'(x)|^p \lambda(\dd x)\right)^{\frac{1}{p}} \nonumber \\
    && = \left(\int_{\G} \frac{1}{1 + \lambda(\G)^{p}} \Big[ 1 + \lambda(\G)^{p}\Big] |f'(x)|^p \lambda(\dd x)\right)^{\frac{1}{p}} \nonumber \\
    && \ge \left(\int_{\G} \frac{1}{1 + \lambda(\G)^{p}} \Big[1 + \lambda(\G)^{p-1}\lambda(\Lambda(x))\Big] |f'(x)|^p \lambda(\dd x)\right)^{\frac{1}{p}} \nonumber \\
    && \ge \left(\int_{\G} \frac{\min(1, \lambda(\G)^{p-1})}{1 + \lambda(\G)^{p}} \Big[ 1 + \lambda(\Lambda(x))\Big] |f'(x)|^p \lambda(\dd x)\right)^{\frac{1}{p}} \nonumber \\
    && = c_1 \norm{f'}_{L^p_{\hat{w}}}, \label{app:eq:lower_bound_S_wLp}
\end{eqnarray}
where we have used the fact that $\norm{f}_{L^p} \ge 0$ to obtain the inequality~\eqref{app:eq:nonnegativeLp}, and recall that $c_1 = \left[ \frac{\min(1, \lambda(\G)^{p-1})}{1 + \lambda(\G)^{p}} \right]^{\frac{1}{p}}$.

Thus, we conclude  from Equations~\eqref{app:eq:upper_bound_S_wLp} and~\eqref{app:eq:lower_bound_S_wLp} that
\[
c_1\norm{f'}_{L^p_{\hat{w}}} \le \norm{f}_{W^{1, p}} \le c_2 \norm{f'}_{L^p_{\hat{w}}}.
\]
The proof is completed.

\end{proof}

\subsection{Proof for Theorem~\ref{thrm:closed_form_regSobolevIPM}}\label{sec:proof_closedform}

\begin{proof}
Consider a critic function $f\in W_0^{1, p'}(\G, \lambda)$. Then by Definition~\ref{def:Sobolev}, we have
\begin{equation}\label{eq:OSF_representation}
f(x) = f(z_0) + \int_{[z_0,x]} f'(y) \lambda(\mathrm{d}y)\quad \mbox{for all}\quad x\in \G.
\end{equation}
Using \eqref{eq:OSF_representation}, leveraging the indicator function of the shortest path $[z_0, x]$, and notice that $\mu(\G) = 1$, we get
\begin{align*}
\int_\G f(x) \mu(\mathrm{d}x) &= \int_\G f(z_0) \mu(\mathrm{d}x) + \int_\G  \int_{[z_0,x]} f'(y) \lambda(\mathrm{d}y) \mu(\mathrm{d}x)\\
&= f(z_0) + \int_\G  \int_{\G}  {\bf{1}}_{[z_0,x]}(y) \, f'(y) \lambda(\mathrm{d}y) \mu(\mathrm{d}x).
\end{align*}
Then, applying Fubini's theorem to interchange the order of integration in the above last integral, we obtain
\begin{align*}
\int_\G f(x) \mu(\mathrm{d}x) &= f(z_0) + \int_\G  \int_{\G}  {\bf{1}}_{[z_0,x]}(y) \, f'(y)  \mu(\mathrm{d}x) \lambda(\mathrm{d}y)\\
&= f(z_0) + \int_\G  \Big(\int_{\G}  {\bf{1}}_{[z_0,x]}(y) \,  \mu(\mathrm{d}x)\Big) f'(y)  \lambda(\mathrm{d}y).
\end{align*}
Using the definition of $\Lambda(y)$ in Equation~\eqref{sub-graph}, we can rewrite it as 
\begin{align*}
\int_\G f(x) \mu(\mathrm{d}x) = f(z_0) + \int_{\G} f'(y)  \mu(\Lambda(y)) \, \lambda(\mathrm{d}y).
\end{align*}

By exactly the same arguments, we also have 
\begin{align*}
\int_\G f(x) \nu(\mathrm{d}x) 
 = f(z_0) + \int_{\G} f'(y)  \nu(\Lambda(y)) \, \lambda(\mathrm{d}y).
\end{align*}

Consequently, the regularized Sobolev IPM in Equation~\eqref{eq:regSobolevIPM} can be reformulated as
\begin{align}\label{eq:reformulation}
\hat{\calS}_{p}(\mu, \nu)  = 
\sup_{f \in \calB(p', \hat{w})} \left| \int_{\G} f'(x) \big[ \mu(\Lambda(x)) -  \nu(\Lambda(x))\big] \, \lambda(\mathrm{d}x) \right|,
\end{align}
where we recall that $\calB(p', \hat{w}) := \left\{f \in W^{1, p'}_0 : \norm{f'}_{L^{p'}_{\hat{w}}} \leq 1 \right\}$ (see Equation~\eqref{eq:weightedLp_ball_regularized}). 

Observe that, we have on one hand
\[
\{ f': \, f \in \calB(p', \hat{w})\} \subset \{g\in L^{p'}(\G, \lambda): \, \|g\|_{L^{p'}_{\hat{w}}}\leq 1  \}.
\]
On the other hand, for any $g\in L^{p'}(\G, \lambda)$, we have $g=f'$ with $f(x) \coloneqq  \int_{[z_0,x]} g(y) \lambda(\mathrm{d}y)\in L^{p'}(\G, \lambda)$. 

Therefore, we conclude that
\begin{equation}\label{app:eq:unitball_func_grad}
\{ f': \, f\in \calB(p', \hat{w})\} = \{g\in L^{p'}(\G, \lambda): \, \|g\|_{L^{p'}_{\hat{w}}} \leq 1  \}.
\end{equation}

Consequently, if we let $\hat f(x) :=  \frac{\mu(\Lambda(x)) -  \nu(\Lambda(x))}{\hat{w}(x)}$ for  $ x \in \G$, then Equation~\eqref{eq:reformulation} can be recasted as 
\begin{align}
\hat{\calS}_{p}(\mu, \nu)  &= 
\sup_{g\in L^{p'}(\G, \lambda): \, \norm{g}_{L^{p'}_{\hat{w}}} \le 1} \left| \int_{\G} \hat{w}(x) \hat f(x) g(x) \lambda(\mathrm{d}x) \right| 
\label{app:eq:dualnorm_regSobolevIPM} 
\\
&= \left( \int_{\G} \hat{w}(x) |\hat f(x)|^p \lambda(\dd x) \right)^{\frac{1}{p}} \label{eq:dualnorm_weightedLp_cont}\\
&= \left( \int_{\G} \hat{w}(x)^{1-p} | \mu(\Lambda(x)) -  \nu(\Lambda(x)) |^p \lambda(\dd x) \right)^{\frac{1}{p}},\nonumber
\end{align}
where  Equation~\eqref{eq:dualnorm_weightedLp_cont} is followed by the dual norm of the weighted $L^{p'}\!(\G, \lambda)$ with the weighting function $\hat{w}$.

Hence, we have
\begin{equation*}
    \hat{\calS}_{p}(\mu, \nu)^p = \int_{\G} \hat{w}(x)^{1-p} \left| \mu(\Lambda(x)) - \nu(\Lambda(x)) \right|^p \lambda(\dd x).
\end{equation*}
The proof is completed.
\end{proof}

\subsection{Proof for Theorem~\ref{thrm:regSobolevIPM_discrete}}\label{app:subsec:proof_thrm:regSobolevIPM_discrete}

\begin{proof}
    We consider the length measure on graph $\G$ for $\lambda$. Thus, we have $\lambda(\{x\}) = 0$ for all $x \in \G$. 
    
    Consequently, we have
    \begin{equation}\label{eq:edge_regSobolevIPM}
    \hat{\calS}_{p}(\mu, \nu)^p = \sum_{e=\langle u,v\rangle\in E}   \int_{(u,v)} \hat{w}(x)^{1-p} \left| \mu(\Lambda(x)) -  \nu(\Lambda(x)) \right|^p \lambda(\dd x).
    \end{equation}
    
    Additionally, we consider input probability measures $\mu, \nu$ supported on nodes in $V$ of graph $\G$. Thus, for all edge $e=\langle u,v\rangle\in E$, and any point $x \in (u, v)$, we have
    \[
    \mu(\Lambda(x)) -  \nu(\Lambda(x))= \mu(\Lambda(x)\setminus (u,v)) -  \nu(\Lambda(x)\setminus (u,v)).
    \]
    
    Hence, we can rewrite Equation~\eqref{eq:edge_regSobolevIPM} as \begin{equation}\label{eq:remove_interior}
    \hat{\calS}_{p}(\mu, \nu)^p = \sum_{e=\langle u,v\rangle\in E}   \int_{(u,v)} \hat{w}(x)^{1-p} \left| \mu(\Lambda(x)\setminus (u,v)) -  \nu(\Lambda(x)\setminus (u,v)) \right|^p \lambda(\dd x).
    \end{equation}

    Let us consider edge $e=\langle u,v\rangle \in E$.  Then for any $x\in (u,v)$, we have $y\in \G\setminus (u,v)$ belongs to $\Lambda(x)$ if and only if $y\in \gamma_e$ where we recall that $\Lambda(x)$ and $\gamma_e$ are defined in Equation~\eqref{sub-graph}. Thus, we have
    \[
    \Lambda(x)\setminus (u,v) =\gamma_e, \qquad \forall x \in (u, v).
    \]
    Using this fact, we can rewrite Equation~\eqref{eq:remove_interior} as
    \begin{equation}\label{eq:remove_interior_edge}
    \hat{\calS}_{p}(\mu, \nu)^p = \sum_{e=\langle u,v\rangle\in E} \left| \mu(\gamma_e) -  \nu(\gamma_e) \right|^p  \int_{(u,v)} \hat{w}(x)^{1-p} \lambda(\dd x).
    \end{equation}

    We next want to compute the integral in \eqref{eq:remove_interior_edge}
    for each edge $\langle u,v\rangle\in E$. For this, recall that $\hat{w}(x) = 1 + \lambda(\Lambda(x))$ (see Equation~\eqref{eq:weighting_func_wLp}). Without loss of generality, assume that $d_{\G}(z_0, u) \le d_{\G}(z_0, v)$, i.e., among two nodes $u, v$ of the edge $e$, node $v$ is farther away from the root node $z_0$ than node $u$. 

    Notice that for $x \in (u, v)$, we can write $x = v + t (u-v)$ for $t \in (0, 1)$. With this change of variable, we have
    \begin{equation*}
    \int_{(u, v)} \left[ 1 + \lambda(\Lambda(x)) \right]^{1-p} \lambda(\dd x) = \int_0^1 \left[ 1 + \lambda(\Lambda( v + t (u-v))) \right]^{1-p}   w_e\dd t
    \end{equation*}

    Moreover, we have
    \[
    \lambda(\Lambda( v + t (u-v))) = \lambda(\Lambda(v)) + \lambda([v,  v + t (u-v)]) = 
    \lambda(\gamma_e) + w_e t.
    \]
    
    Therefore, 
    \begin{equation*}
    \int_{(u, v)} \left[ 1 + \lambda(\Lambda(x)) \right]^{1-p} \lambda(\dd x) = \int_0^1 \left[ 1 + \lambda(\gamma_e) + w_e t \right]^{1-p} w_e\dd t.
    \end{equation*}
    The last integral can be computed easily depending on the case $p=2$ or $p\neq 2$.
    As a consequence, we obtain 
\begin{equation*}\label{eq:remove_interior_edge_weight}
    \int_{(u, v)} \left[ 1 +  \lambda(\Lambda(x)) \right]^{1-p} \lambda(\dd x) = \left\{
    \begin{array}{ll}
          \log{\left(1 + \frac{w_e}{1 + \lambda(\gamma_e)} \right)} & p = 2 \\
          \frac{ \left(1 + \lambda(\gamma_e)+ w_e \right)^{2-p} - \left(1 + \lambda(\gamma_e) \right)^{2-p} }{2-p} & \text{otherwise.}
    \end{array} 
    \right. 
    \end{equation*}
Thus, we have $\int_{(u, v)} \left[ 1 + \lambda(\Lambda(x)) \right]^{1-p} \lambda(\dd x) = \beta_e$ (see Equation~\eqref{eq:closed_form_edge_weight_regSobolevIPM}).
This  together with Equation~\eqref{eq:remove_interior_edge} yields
\begin{equation}\label{eq:closed_form_discrete}
    \hat{\calS}_{p}(\mu, \nu) = \left( \sum_{e\in E} \beta_e | \mu(\gamma_e) -  \nu(\gamma_e) |^p \right)^{\frac{1}{p}}.
\end{equation}
 
Hence,  the proof is  completed. 

\end{proof}

\subsection{Proof for Theorem~\ref{thrm:metrize}}\label{app:subsec:proof_thrm:metrize}

\begin{proof}

For $1 \le p \le \infty$, we will prove that the regularized Sobolev IPM $\hat{\calS}_{p}(\cdot,\cdot)$ satisfies: (i) nonnegativity, (ii) indiscernibility, (iii) symmetry, and (iv) triangle inequality.


\textbf{(i) Nonnegativity.} By choosing $f=0$ in Definition~\ref{def:regSobolevIPM}, we see that $\hat{\calS}_{p}(\mu,\nu) \ge 0$ for every  measures $(\mu, \nu)$ in $\calP(\G) \times \calP(\G)$. 

Therefore, the regularized Sobolev IPM $\hat{\calS}_p$ is nonnegative.


\textbf{(ii) Indiscernibility.} Assume that $\hat{\calS}_p(\mu,\nu ) =0$. Then we must have 
\begin{align}\label{int_identity}
\int_\G f(x) \mu(\mathrm{d}x) - \int_\G f(x) \nu(\mathrm{d}x) =0
\end{align}
for all $f\in W_0^{1,p'}(\G, \lambda)$ satisfying the constraint $\|f'\|_{L^{p'}_{\hat{w}}}\leq 1$. 
Indeed, if by contradiction that \eqref{int_identity} is not true, then due to $\hat{\calS}_p(\mu,\nu ) = 0$ there exists a function $g \in W_0^{1,p'}(\G, \lambda)$ such that $\|g'\|_{L^{p'}_{\hat{w}}}\leq 1$, and $\int_\G g(x) \mu(\mathrm{d}x) - \int_\G g(x) \nu(\mathrm{d}x) < 0$.
Then, by choosing critic function $f=-g$ in Definition~\ref{def:regSobolevIPM}, we see that $\hat{\calS}_p(\mu,\nu ) > 0$. This contradicts the assumption $\hat{\calS}_p(\mu,\nu ) = 0$. 

Therefore, Equation~\eqref{int_identity} holds true. Consequently, we have 
\[
\int_\G f(x) \mu(\mathrm{d}x) = \int_\G f(x) \nu(\mathrm{d}x), 
\]
for every $f\in W_0^{1,p'}(\G, \lambda)$ with $\|f'\|_{L^{p'}_{\hat{w}}}\leq 1$. This gives $\mu = \nu$ as desired.

\textbf{(iii) Symmetry.} Observe that if $f\in W_0^{1,p'}(\G, \lambda)$ with $\|f'\|_{L^{p'}_{\hat{w}}}\leq 1$, then we have $(-f) \in W_0^{1,p'}(\G, \lambda)$ with $\|-f'\|_{L^{p'}_{\hat{w}}} = \|f'\|_{L^{p'}_{\hat{w}}} \leq 1$. As a consequence, we have
\[
\hat{\calS}_p(\mu,\nu ) = \hat{\calS}_p(\nu,\mu).
\]

\textbf{(iv) Triangle inequality.} Let $\mu,\nu,\sigma$ be probability measures in $\calP(\G)$. Then, for any critic function $f\in W_0^{1,p'}(\G, \lambda)$ satisfying $\|f'\|_{L^{p'}_{\hat{w}}}\leq 1$, we have
\begin{align*}
\left| \int_\G f(x) \mu(\mathrm{d}x) - \int_\G f(x) \nu(\mathrm{d}x) \right| &= \left| \left( \int_\G f(x) \mu(\mathrm{d}x) - \int_\G f(x) \sigma(\mathrm{d}x) \right)  + \left( \int_\G f(x) \sigma(\mathrm{d}x) - \int_\G f(x) \nu(\mathrm{d}x)\right) \right|\\
&\leq \left| \int_\G f(x) \mu(\mathrm{d}x) - \int_\G f(x) \sigma(\mathrm{d}x) \right|  + \left| \int_\G f(x) \sigma(\mathrm{d}x) - \int_\G f(x) \nu(\mathrm{d}x) \right| \\
&\leq \hat{\calS}_p(\mu,\sigma ) + \hat{\calS}_p(\sigma,\nu ).
\end{align*}
By taking the supremum over critic function $f$, this implies that 
\[
\hat{\calS}_{p}(\mu, \nu) \leq \hat{\calS}_{p}(\mu,\sigma) + \hat{\calS}_{p}(\sigma, \nu).
\]

Hence, from these above properties, we conclude that the regularized Sobolev IPM $\hat{\calS}_{p}(\mu,\nu)$ is a metric for probability measures on the space $\calP(\G)$.

The proof is completed.

\end{proof}

\subsection{Proof for Theorem~\ref{thrm:relationSobolevIPM}}\label{app:subsec:proof_thrm:relationSobolevIPM}

\begin{proof}


Given a positive number $c > 0$, let us consider $\calB(p, \hat{w}, c) := \left\{ f \in W_0^{1, p}(\G, \lambda) : \norm{f'}_{L^p_{\hat{w}}} \le \frac{1}{c} \right\}$.  We define the IPM distance w.r.t. $\calB(p, \hat{w}, c)$ as follows
\begin{equation}\label{eq:IPM_tmp}
\bar{\calS}_{p, c}(\mu, \nu) := \sup_{f \in \calB(p', \hat{w}, c)} \left| \int_{\G} f(x) \mu(\text{d}x) - \int_{\G} f(y) \nu(\text{d}y) \right|.
\end{equation}
By exploiting the graph structure for the IPM objective function
and applying a similar reasoning as in the proof of identity \eqref{eq:reformulation} in \S\ref{sec:proof_closedform}, we can rewrite Equation~\eqref{eq:IPM_tmp} as 
\begin{align}\label{eq:reformulation_IMP_tmp}
\bar{\calS}_{p, c}(\mu, \nu)  = 
\sup_{f \in \calB(p', \hat{w}, c)} \left| \int_{\G} f'(x) \big[ \mu(\Lambda(x)) -  \nu(\Lambda(x))\big] \, \lambda(\mathrm{d}x) \right|.
\end{align}
Additionally, by using a similar reasoning as in the proof of \eqref{app:eq:unitball_func_grad} in \S\ref{sec:proof_closedform},  we have
\begin{equation}\label{app:eq:unitball_func_grad_IPM_tmp}
\left\{ f': \, f\in \calB(p', \hat{w}, c)\right\} = \left\{g\in L^{p'}(\G, \lambda): \, \|g\|_{L^{p'}_{\hat{w}}} \leq \frac{1}{c}  \right\}.
\end{equation}

Let $\bar f(x) :=  \frac{\mu(\Lambda(x)) -  \nu(\Lambda(x))}{c \, \hat{w}(x)}$ for $ x \in \G$. Then by using \eqref{app:eq:unitball_func_grad_IPM_tmp}, Equation~\eqref{eq:reformulation_IMP_tmp} can be recasted as 
\begin{align}
\bar{\calS}_{p, c}(\mu, \nu)  &= 
\sup_{g\in L^{p'}(\G, \lambda):\, \norm{g}_{L^{p'}_{\hat{w}}} \le \frac{1}{c}} \left| \int_{\G} \hat{w}(x) \bar f(x) [c \, g(x)] \lambda(\mathrm{d}x) \right| \nonumber \\
&= 
\sup_{\bar g\in L^{p'}(\G, \lambda):\, \norm{\bar g}_{L^{p'}_{\hat{w}}} \le 1} \left| \int_{\G} \hat{w}(x) \bar f(x) \bar g(x)\lambda(\mathrm{d}x) \right| \nonumber \\
&= \left( \int_{\G} \hat{w}(x) |\bar f(x)|^p \lambda(\dd x) \right)^{\frac{1}{p}} \label{eq:dualnorm_weightedLp}\\
&= \left( \int_{\G} \frac{1}{c^p} \hat{w}(x)^{1-p} | \mu(\Lambda(x)) -  \nu(\Lambda(x)) |^p \lambda(\dd x) \right)^{\frac{1}{p}} \\
&= \frac{1}{c} \hat{\calS}_p(\mu, \nu). \label{eq:relation_regSobolevIPM_IPMtmp}
\end{align}
where the Equation~\eqref{eq:dualnorm_weightedLp} is followed by the dual norm of the weighted $L^{p'}\!(\G, \lambda)$ with weighting function $\hat{w}$; and Equation~\eqref{eq:relation_regSobolevIPM_IPMtmp} is followed by the closed-form expression of the regularized Sobolev IPM in Theorem~\ref{thrm:closed_form_regSobolevIPM}.

Notice that for $1 \le p < \infty$, from Theorem~\ref{thrm:S_wLp_norm}, we have 
\[
c_1 \norm{f'}_{L^{p'}_{\hat{w}}} \le \norm{f}_{W^{1, p'}} \le c_2 \norm{f'}_{L^{p'}_{\hat{w}}}.
\]
This implies that
\[
\calB(p', \hat{w}, c_1) \supseteq \calB(p') \supseteq \calB(p', \hat{w}, c_2).
\]
Therefore, for probability measures $\mu, \nu \in \calP(\G)$, we have
\begin{equation}\label{eq:relation_SobolevIPM_IPMtmp}
\bar{\calS}_{p, c_1}(\mu, \nu) \ge \calS_p(\mu, \nu) \ge \bar{\calS}_{p, c_2}(\mu, \nu).
\end{equation}

It follows from Equations~\eqref{eq:relation_regSobolevIPM_IPMtmp} and~\eqref{eq:relation_SobolevIPM_IPMtmp} that
\begin{equation*}\label{app:eq:lowerbound_reg_SobolevIPM}
\frac{1}{c_1} \, \hat{\calS}_p(\mu, \nu) \ge \calS_p(\mu, \nu) \ge \frac{1}{c_2} \, \hat{\calS}_p(\mu, \nu),
\end{equation*}
which gives
\begin{equation*}
c_1 \, \calS_p(\mu, \nu) \le \hat{\calS}_p(\mu, \nu) \le c_2 \, \calS_p(\mu, \nu).
\end{equation*}

The proof is complete.

\end{proof}

\subsection{Proof for Proposition~\ref{prop:relationSobolevTransport}}\label{app:subsec:proof_ST_relation}

\begin{proof}

Observe that $0 \le \lambda(\Lambda(x)) \le \lambda(\G)$ for all $x \in \G$. Consequently, we have
\begin{equation*}
\frac{1}{1 + \lambda({\G})} (1 + \lambda(\Lambda(x))) \le 1 \le 1 + \lambda(\Lambda(x)), \quad \forall x \in \G.
\end{equation*}
Therefore, for $1 \le p < \infty$, we have
\begin{equation}\label{app:weight_relation}
(1 + \lambda({\G}))^{p-1} \hat{w}(x)^{1-p} \ge 1 \ge \hat{w}(x)^{1-p}, \quad \forall x \in \G.
\end{equation}
Additionally, recall that the Sobolev transport $\mathcal{ST}_{\!p}$ for $\mu, \nu \in \calP(\G)$ admits a closed-form expression~\citep[Proposition 3.5]{le2022st}
\begin{equation}\label{app:STp-closed-form-proof}
\mathcal{ST}_{\!p}(\mu, \nu) = \left( \int_{\G} \left| \mu(\Lambda(x)) - \nu(\Lambda(x)) \right|^p \lambda(\dd x) \right)^{\frac{1}{p}}.
\end{equation}
By combining the closed-form expression of the regularized Sobolev IPM (Equation~\eqref{eq:closed_form_regSobolevIPM} in Theorem~\ref{thrm:closed_form_regSobolevIPM}) with Equations~\eqref{app:weight_relation} and \eqref{app:STp-closed-form-proof}, we obtain
\begin{equation*}
\hat{\calS}_p(\mu, \nu) \le \mathcal{ST}_{\!p}(\mu, \nu) \le (1 + \lambda({\G}))^{\frac{p-1}{p}} \hat{\calS}_p(\mu, \nu).
\end{equation*}
It follows that
\begin{equation}
(1 + \lambda({\G}))^{\frac{1-p}{p}} \mathcal{ST}_{\!p}(\mu, \nu) \le \hat{\calS}_p(\mu, \nu) \le \mathcal{ST}_{\!p}(\mu, \nu).
\end{equation}

The proof is complete.

\end{proof}



\subsection{Proof for Proposition~\ref{prop:upper}}\label{app:subsec:proof-different-pq}

\begin{proof}
For convenience, let $\bar{h}(x) : = \mu(\Lambda(x)) - \nu(\Lambda(x))$. Recall also that $p' = \frac{p}{p-1}$ and $q' = \frac{q}{q-1}$ are respectively the conjugate of $p$ and $q$. Then
it follows from Theorem~\ref{thrm:closed_form_regSobolevIPM} that
\begin{equation*}
    \hat{\calS}_{p}(\mu, \nu)^p = \int_{\G} \hspace{-0.3em} \hat{w}(x)^{1-p} |\bar{h}(x)|^p \lambda(\dd x)
     =  \int_{\G} \hspace{-0.3em} \left(\hat{w}(x)^{\frac{q-p}{q}} \right) \left( \hat{w}(x)^{\frac{p(1-q)}{q}} |\bar{h}(x)|^p\right) \lambda(\dd x).
\end{equation*}
We can apply H\"older inequality with $l = \frac{q}{p}>1$ and $l' = \frac{q}{q-p}$ to obtain
\begin{align*}
    \hat{\calS}_{p}(\mu, \nu)^p 
    &\leq 
  \left[\int_{\G} \hspace{-0.3em} \left( \hat{w}(x)^{\frac{q-p}{q}} \right)^{\frac{q}{q-p}} \lambda(\dd x)\right]^{\frac{q-p}{q}}
  \left[  \int_{\G} \hspace{-0.3em} \left(\hat{w}(x)^{\frac{p(1-q)}{q}} |\bar{h}(x)|^p\right)^{\frac{q}{p}} \lambda(\dd x)\right]^{\frac{p}{q}} \\
  &= \left[\int_{\G} \hspace{-0.3em} \hat{w}(x) \lambda(\dd x)\right]^{\frac{q-p}{q}}
  \left[  \int_{\G} \hspace{-0.3em} \hat{w}(x)^{1-q} |\bar{h}(x)|^q  \lambda(\dd x)\right]^{\frac{p}{q}}.
\end{align*}

In addition, it follow directly from the definition in Equation \eqref{eq:weighting_func_wLp} that 
\[
\hat{w}(x) \leq 1 +\lambda(\G).
\]
for every $x\in\G$. Therefore, we further deduce from above estimate and Theorem~\ref{thrm:closed_form_regSobolevIPM} that
\begin{align*}
    \hat{\calS}_{p}(\mu, \nu)^p 
    &\leq 
  \left[\int_{\G} \hspace{-0.3em} \left(1 +\lambda(\G) \right) \lambda(\dd x)\right]^{\frac{q-p}{q}}
  \left[  \int_{\G} \hspace{-0.3em} \hat{w}(x)^{1-q} |\bar{h}(x)|^q  \lambda(\dd x)\right]^{\frac{p}{q}}\\
  & = \left[ \lambda(\G) \left(1 +\lambda(\G) \right) \right]^{\frac{q-p}{q}}   \hat{\calS}_{q}(\mu, \nu)^p.
\end{align*}
Hence, we obtain
\begin{equation*}
\hat{\calS}_p(\mu, \nu) \le
\left[ \lambda(\G) \left(1 +\lambda(\G) \right) \right]^{\frac{1}{p} - \frac{1}{q}}   \hat{\calS}_{q}(\mu, \nu).
\end{equation*}
This completes the proof of the proposition.
\end{proof}

\subsection{Proof for Proposition~\ref{prop:tree}}\label{app:subsec:proof-relation-regSIPM1-W1}

\begin{proof}
From Proposition~\ref{prop:relationSobolevTransport}, we have
    \begin{equation*}
    \hat{\calS}_1(\mu, \nu) = \mathcal{ST}_{\!1}(\mu, \nu).
    \end{equation*}
Additionally, following~\citet[Corollary 4.3]{le2022st}, when graph $\G$ is a tree and $\lambda$ is the length measure on $\G$, we have
\[
\mathcal{ST}_{\!1}(\mu, \nu) = \calW_1(\mu, \nu),
\]
where  the $1$-Wasserstein distance $\calW_1$
 is defined w.r.t. the ground cost $d_{\G}$.

Hence, we obtain
\[
\hat{\calS}_1(\mu, \nu) = \calW_1(\mu, \nu).
\]
The proof is complete.
\end{proof}

\subsection{Proof for Proposition~\ref{prop:w1-vs-sp}}\label{app:subsec:proof-relation-regSIPMp-W1}

\begin{proof}
    From Proposition~\ref{prop:tree}, we have
    \[
    \hat{\calS}_1(\mu, \nu) = \calW_1(\mu, \nu).
    \]
    Additionally,  Proposition~\ref{prop:upper} gives
    \begin{eqnarray*}
        \hat{\calS}_1(\mu, \nu) \leq  \left[ \lambda(\G)(1 + \lambda(\G)) \right]^{\frac{p-1}{p}}\hat\calS_p(\mu,\nu ).
    \end{eqnarray*}
    Therefore, we obtain
    \[
    \calW_1(\mu, \nu) \leq \left[ \lambda(\G)(1 + \lambda(\G)) \right]^{\frac{p-1}{p}}\hat\calS_p(\mu,\nu ).
    \]
   It then follows that
    \[
    \hat\calS_p(\mu,\nu ) \geq \left[ \lambda(\G)(1 + \lambda(\G)) \right]^{\frac{1-p}{p}} \calW_1(\mu, \nu).
    \]
    The proof is complete.
    
\end{proof}

\subsection{Proof for Proposition~\ref{prop:neg_def}}\label{app:subsec:proof-negdef}

\begin{proof}

For $x, z \in \mathbb{R}^{m}$, consider the following function 
\[
\ell_p(x,z) = \norm{x-z}_p = \left(\sum_{i=1}^m \left|x_{(i)} - z_{(i)} \right|^p\right)^{1/p},
\]
where $x_{(i)}$ is the $i^{th}$ coordinate of $x$.  We will prove that for $1 \le p \le 2$, the functions $\ell_p$ and $\ell_p^p$ are negative definite.

For $a, b \in \mathbb{R}$, it is obvious that the function $(a, b) \mapsto (a - b)^2$ is negative definite. For $1 \le p \le 2$, following \citet[Corollary~2.10, pp.78]{Berg84}, the function $(a, b) \mapsto |a - b|^p$ is also negative definite. 

Consequently, the function $\ell_p^p$ is negative definite since it is a sum of negative definite functions. Moreover, following \citet[Corollary~2.10, pp.78]{Berg84}, we also have that the function $\ell_p$ is  negative definite. 

Let $m$ be the number of edges in the graph $\G$, i.e., $|E| = m$. Following Theorem~\ref{thrm:regSobolevIPM_discrete}, we regard $({\beta}_e)^{\frac{1}{p}} \mu(\gamma_e)$ for each edge $e \in E$ of graph $\G$ as a feature map for probability measure $\mu$ onto $\R^{m}_{+}$. Therefore, $\hat{\calS}_p$ is equal to the function $\ell_p$ between these feature maps.

Hence, $\hat{\calS}_p$ and $\hat{\calS}_p^p$ are negative definite for $1 \le p \le 2$. The proof is complete.
\end{proof}

\subsection{Proof for Proposition~\ref{prop:divisibility}}

\begin{proof}

For probability measures $\mu, \nu$ supported on nodes in $V$ of graph $\G$, $1 \le p \le 2$, and an integer $\tau \in \mathbb{N}^{*}$, we define the following kernels
\begin{eqnarray}
k_{\hat{\calS}_p}^{\tau}(\mu, \nu) := 
 \exp\left(-t \frac{\hat{\calS}_p(\mu, \nu)}{\tau}\right),\\
 k_{\hat{\calS}_p^p}^{\tau}(\mu, \nu) := 
 \exp\left(-t \frac{\hat{\calS}_p(\mu, \nu)^p}{\tau}\right).
\end{eqnarray}
 Observe that $k_{\hat{\calS}_p}^{\tau}(\mu, \nu)^{\tau} = k_{\hat{\calS}_p}(\mu, \nu)$ and $k_{\hat{\calS}_p^p}^{\tau}(\mu, \nu)^{\tau} = k_{\hat{\calS}_p^p}(\mu, \nu)$.

 Furthermore, both kernels $k_{\hat{\calS}_p}^{\tau}$ and $k_{\hat{\calS}_p^p}^{\tau}$ are positive definite.
 
Hence, following~\citet[\S3, Definition~2.6, pp.~76]{Berg84}, the kernels $k_{\hat\calS_p}$ and $k_{\hat\calS^p_p}$ are infinitely divisible. The proof is complete.

\end{proof}

\section{Reviews}\label{app:sec:reviews}

In this section, we give a review for related notions used in the development of our proposed approach.

\subsection{A Review on Functional Spaces}\label{app:sec:FuncSpaces}

We give a review on the $L^p$ space, the weighted $L^p$ space, and Sobolev norm including the case the exponent $p = \infty$.

\paragraph{$L^{p}$ space.} For a nonnegative Borel measure $\lambda$ on $\G$, denote $L^p( \G, \lambda)$ as the space of all Borel measurable functions $f:\G\to \R$ such that $\int_\G |f(y)|^p \lambda(\mathrm{d}y) <\infty$. For $p=\infty$, we instead assume that $f$ is bounded $\lambda$-a.e. Functions $f_1, f_2 \in L^p( \G, \lambda)$ are considered to be the same if $f_1(x) =f_2(x)$ for $\lambda$-a.e. $x\in\G$. 

Then, $L^p( \G, \lambda)$ is a normed space with the norm defined by
\[
\|f\|_{L^p} := \left(\int_\G |f(y)|^p \lambda(\dd y)\right)^\frac{1}{p} \text{ for } 1\leq p < \infty.
\]
On the other hand, for $p = \infty$ we have
\[
\|f\|_{L^{\infty}} := \inf\left\{t \in \R:\, |f(x)|\leq t \mbox{ for $\lambda$-a.e. } x\in\G\right\}.
\]

\paragraph{${L^{p}_{\hat{w}}}$ space.} For a nonnegative Borel measure $\lambda$ on $\G$, and a positive weight function $\hat{w}$ on $G$, i.e., $\hat{w}(x) > 0, \forall x \in \G$, denote $L^p_{\hat{w}}( \G, \lambda)$ as the space of all Borel measurable functions $f:\G\to \R$ such that $\int_\G \hat{w}(x) |f(x)|^p \lambda(\mathrm{d}x) <\infty$. For $p=\infty$, we instead assume that $f$ is bounded $\hat{w}\lambda$-a.e. Then, $L^p_{\hat{w}}( \G, \lambda)$ is a normed space with the norm defined by
\[
\|f\|_{L^p_{\hat{w}}} := \left(\int_\G \hat{w}(x) |f(x)|^p \lambda(\dd x)\right)^\frac{1}{p} \text{ for } 1\leq p < \infty.
\]
For the case $p = \infty$, as $\hat{w}(x) > 0$ for every $x \in \G$ we have
\begin{align*}
\|f\|_{L^\infty_{\hat{w}}} 
&:= \inf\left\{t \in \R:\, |f(x)|\leq t \mbox{ for $(\hat{w}\lambda)$-a.e. } x\in\G\right\}\\
&=\inf\left\{t \in \R:\, |f(x)|\leq t \mbox{ for $\lambda$-a.e. } x\in\G\right\}\\
&=\|f\|_{L^{\infty}} .
\end{align*}

\paragraph{Sobolev norm.} For $1 \le p \le \infty$ and a function $f \in W^{1,p}(\G, \lambda)$, its Sobolev norm~\citep[\S3.1]{adams2003sobolev} is
\begin{equation}\label{app:eq:SobolevNorm}
    \norm{f}_{W^{1,p}} = \left( \norm{f}_p^p + \norm{f'}_p^p \right)^{\frac{1}{p}}.
\end{equation}
When $p=\infty$, its norm is defined as follows:
\begin{equation}\label{app:eq:SobolevNorm_Infty}
    \norm{f}_{W^{1,\infty}} =  \norm{f}_{\infty} + \norm{f'}_{\infty}.
\end{equation}


\subsection{A Review on Sobolev Transport}\label{app:sec:ST}\label{app:subsec:review-ST}

We give a review on the Sobolev transport for probability measures on a graph, and the length measure on a graph.

\textbf{Sobolev transport.} For $\mu, \nu \in \calP(\G)$, and $1 \le p \le \infty$, the $p$-order Sobolev transport (ST)~\citep[Definition 3.2]{le2022st} is defined as 
\begin{equation} \label{eq:distance}
\mathcal{ST}_p(\mu,\nu ) \coloneqq  \left\{
\begin{array}{cl}
 \sup \Big[\int_\G f(x) \mu(\mathrm{d}x) - \int_\G f(x) \nu(\mathrm{d}x)\Big] \\
 \hspace{0.8em} \mathrm{s.t.} \, f  \in W^{1,p'} \hspace{-0.3em} (\G, \lambda),  \, \|f'\|_{L^{p'} (\G, \lambda)}\leq 1,
\end{array}
\right.
\end{equation}
where we write $f'$ for the generalized graph derivative of $f$, and $W^{1,p'} \hspace{-0.3em} (\G, \lambda)$ for the graph-based Sobolev space on $\G$.

\begin{proposition}[Closed-form expression of Sobolev transport~\citep{le2022st}]\label{app:prop:closed-form_ST}
Let $\lambda$ be any nonnegative Borel measure on $\G$, and let  $1\leq p < \infty$. Then, we have 
\[
\mathcal{ST}_{p}(\mu,\nu )
= \left( \int_{\G} | \mu(\Lambda(x)) -  \nu(\Lambda(x))|^p \, \lambda(\mathrm{d}x) \right)^{\frac{1}{p}},
\]
where $\Lambda(x)$ is the subset of $\G$  defined by Equation~\eqref{sub-graph}. 
\end{proposition}

\begin{definition}[Length measure~\citep{le2022st}] \label{def:measure} 
Let $ \lambda^*$ be the unique Borel measure on $\G$ such that the restriction of $\lambda^*$ on any edge is the length measure of that edge. That is, $\lambda^*$  satisfies:
\begin{enumerate}
\item[i)] For  any edge $e$ connecting two nodes $u$ and $v$, we have 
 $\lambda^*(\langle x,y\rangle) = (t-s) w_e$ 
 whenever $x = (1-s) u + s v$ and $y = (1-t)u + t v$ for $s,t \in [0,1)$ with $s \leq t$. Here, recall that $\langle x,y\rangle$ is the line segment in $e$ connecting $x$ and $y$.
 \item[ii)] For any Borel set $F \subset \G$, we have
 \[
 \lambda^*(F) = \sum_{e\in E} \lambda^*(F\cap e).
 \]
\end{enumerate}
\end{definition}

\begin{lemma}[$\lambda^*$ is the length measure on graph~\citep{le2022st}] \label{lem:length-measure}
Suppose that $\G$ has no short cuts, namely, any edge $e$ is a shortest path connecting its two end-points. Then, $\lambda^*$ is a length measure in the sense that
\[
\lambda^*([x,y]) = d_\G(x,y)
\]
for  any  shortest path   $[x,y]$ connecting $x, y$. Particularly, $\lambda^*$ has no atom in the sense that $\lambda^*(\{x\})=0$ for every $x \in \G$. 
\end{lemma}

\subsection{A Review on IPM and Wasserstein Distance}\label{app:subsec:review-IPM-OT}

We give a review on IPM and Wasserstein distance for probability measures.

\paragraph{IPM.} Integral probability metrics (IPM) for probability measures $\mu, \nu$ are defined as follows:

\begin{equation}
    \gamma_{\calF}(\mu, \nu) = \sup_{f \in \calF} \left| \int_{\Omega} f(x) \mu(\text{d}x) - \int_{\Omega} f(y) \nu(\text{d}y)\right|.
\end{equation}

\paragraph{Special case: $1$-Wasserstein distance (dual formulation).} The $1$-Wasserstein distance is a special case of IPM. In particular, for $\calF = \calF_W := \left\{f : \left| f(x) - f(y) \right| \le d_{\G}(x, y) \right\}$ where recall that $d_{\G}$ is the graph metric on graph $\G$, then IPM is equal to the $1$-Wasserstein distance with ground cost $d_{\G}$
\begin{equation}
    \calW(\mu, \nu) = \sup_{f \in \calF_W} \left| \int_{\G} f(x) \mu(\text{d}x) - \int_{\G} f(y) \nu(\text{d}y)\right|.
\end{equation}

Besides Sobolev IPM, and $1$-Wasserstein distance, some other popular instances of IPM include: (i) Besov IPM, (ii) Dudley metric, (iii) total variation metric, (iv) Kolmogorov metric, (v) maximum mean discrepancies (MMD)~\citep{muller1997integral, sriperumbudur2009integral}.

\paragraph{$p$-Wasserstein distance (primal formulation).}
Let $1\leq p <\infty$, suppose that  $\mu$ and $\nu$ are  two nonnegative Borel measures on $\G$ satisfying $\mu(\G) =\nu(\G)=1$. Then, the $p$-Wasserstein distance between $\mu$ and $\nu$ is defined as follows:
\begin{align*}
\calW_p(\mu,\nu)^p 
&= \inf_{\pi \in \Pi(\mu,\nu)}\int_{\G\times\G} d_\G(x,y)^p \pi(\dd x, \dd y),
\end{align*}
where $\Pi(\mu,\nu) := \Big\{ \pi \in \calP(\G \times \G): \, \pi_1= \mu, \, \pi_2= \nu \Big\}$; $\pi_1, \pi_2$ are the first and second marginals of $\pi$ respectively.

\paragraph{$\bullet$ Tree topology.} When graph $\G$ is a tree, the $1$-Wasserstein distance yields a closed-form expression for a fast computation~\citep{LYFC} (i.e., tree-Wasserstein (TW)).~\citet{le2021ept} extend it for unbalanced input measures (i.e., which may have different total masses), and propose a novel regularization approach, named \emph{regularized entropy partial transport} which also admits a closed-form formula for a fast computation. Furthermore,~\citet{tran2025spherical, tran2025distancebased, tran2025geometric, tran2025nonlinear} have recently developed tree-sliced-Wasserstein on tree-structured systems of lines (i.e., tree systems) to extend TW for applications with dynamic-support measures including generative models, and diffusion models.



\subsection{A Review on Kernels}\label{app:subsec:review-kernels} 

We review definitions and theorems about kernels used in our work.

\paragraph{Positive Definite Kernels~\citep[pp.~66--67]{Berg84}.} A kernel function $k: \Omega \times \Omega \rightarrow \R$ is positive definite if for every positive integer $m \ge 2$ and every points $x_1, x_2, ..., x_m \in \Omega$, we have 
\[
\sum_{i, j=1}^m c_i c_j k(x_i, x_j) \ge 0 \qquad \forall c_1,...,c_m \in \R.
\]

\paragraph{Negative Definite Kernels \citep[pp.~66--67]{Berg84}.} A kernel function $k: \Omega \times \Omega \rightarrow \R$ is negative definite if for every integer  $ m \ge 2$ and every points $x_1, x_2, ..., x_m \in \Omega$, we have 
\[
\sum_{i, j=1}^m c_i c_j k(x_i, x_j) \le 0, \qquad \quad \forall c_1,...,c_m \in \R\,\, \text{ s.t. } \, \sum_{i=1}^m c_i = 0.
\]

\paragraph{Theorem 3.2.2 in \citet[pp.~74]{Berg84}.}
Let  $\bar{k}$ be a negative definite kernel function. Then, for every $t>0$, the  kernel 
\[
k(x, z) \Let \exp{\left(- t \bar{k}(x, z)\right)}
\]
is positive definite.

\paragraph{Definition 2.6 in \citet[pp.~76]{Berg84}.}
A positive definite kernel $\bar{k}$ is infinitely divisible if for each $n \in {\N}^{*}$, there exists a positive definite kernel $\bar{k}_n$ such that 
\[
\bar{k} = (\bar{k}_n)^n.
\]

\paragraph{Corollary 2.10 in \citet[pp.~78]{Berg84}.}
Let  $\bar{k}$ be a negative definite kernel function. Then, for $0 < t < 1$, the  kernel 
\[
k(x, z) \Let \left[\bar{k}(x, z)\right]^{t}
\]
is negative definite.

\subsection{Persistence Diagrams}

We refer the readers to \citet[\S2]{kusano2017kernel} for the review on mathematical framework for persistence diagrams (e.g., persistence diagrams, filtrations, persistent homology).


\section{Further Results and Discussions}\label{app:sec:further_results}

In this section, we give further experimental results and further discussions for the regularized Sobolev IPM.

\subsection{Further Results}\label{app:sec:further_results:theory}

\paragraph{For the case $p = \infty$.} The closed-form expression of regularized Sobolev IPM for $p = \infty$ is 
\[
\hat{\calS}_{\infty}(\mu, \nu) = \inf \left\{ t \in \mathcal{R} : \left| \frac{\mu(\Lambda(x)) - \nu(\Lambda(x))}{\hat{w}(x)} \right| \le t \mbox{ for $(\hat{w}\lambda)$-a.e. } x\in\G \right\}.
\]

\begin{proof}
We follow the same reasoning for the proof of Theorem~\ref{thrm:closed_form_regSobolevIPM}. We recall Equation~\eqref{app:eq:dualnorm_regSobolevIPM}
\begin{equation}
\hat{\calS}_{p}(\mu, \nu)  = 
\sup_{\norm{g}_{L^{p'}_{\hat{w}}} \le 1}~\int_{\G} \hat{w}(x) \hat f(x) g(x) \lambda(\mathrm{d}x).
\end{equation}
Therefore, for $p = \infty$, by using the dual norm, and let $\hat{f}(x) := \frac{\mu(\Lambda(x)) - \nu(\Lambda(x))}{\hat{w}(x)} $, for all $x \in \G$, we have 
\begin{equation}
\hat{\calS}_{\infty}(\mu, \nu)  =  \norm{\hat f(x)}_{L^{\infty}_{\hat{w}}},
\end{equation}
which is the weighted $L^{\infty}(\G, \lambda)$-norm with the weighting function $\hat{w}$.

Hence, we obtain
\[
\hat{\calS}_{\infty}(\mu, \nu) = \inf \left\{ t \in \mathcal{R} : \left| \frac{\mu(\Lambda(x)) - \nu(\Lambda(x))}{\hat{w}(x)} \right| \le t \mbox{ for $(\hat{w}\lambda)$-a.e. } x\in\G \right\}.
\]

The proof is completed.
\end{proof}

\subsection{Further Experimental Results}\label{app:sec:further_results:exp}

We present further results for the regularized Sobolev IPM for document classification and TDA for both graphs $\G_{\text{Sqrt}}$ and $\G_{\text{Log}}$ with different numbers of nodes $M$.

\begin{figure*}[ht]
  \begin{center}
    \includegraphics[width=0.75\textwidth]{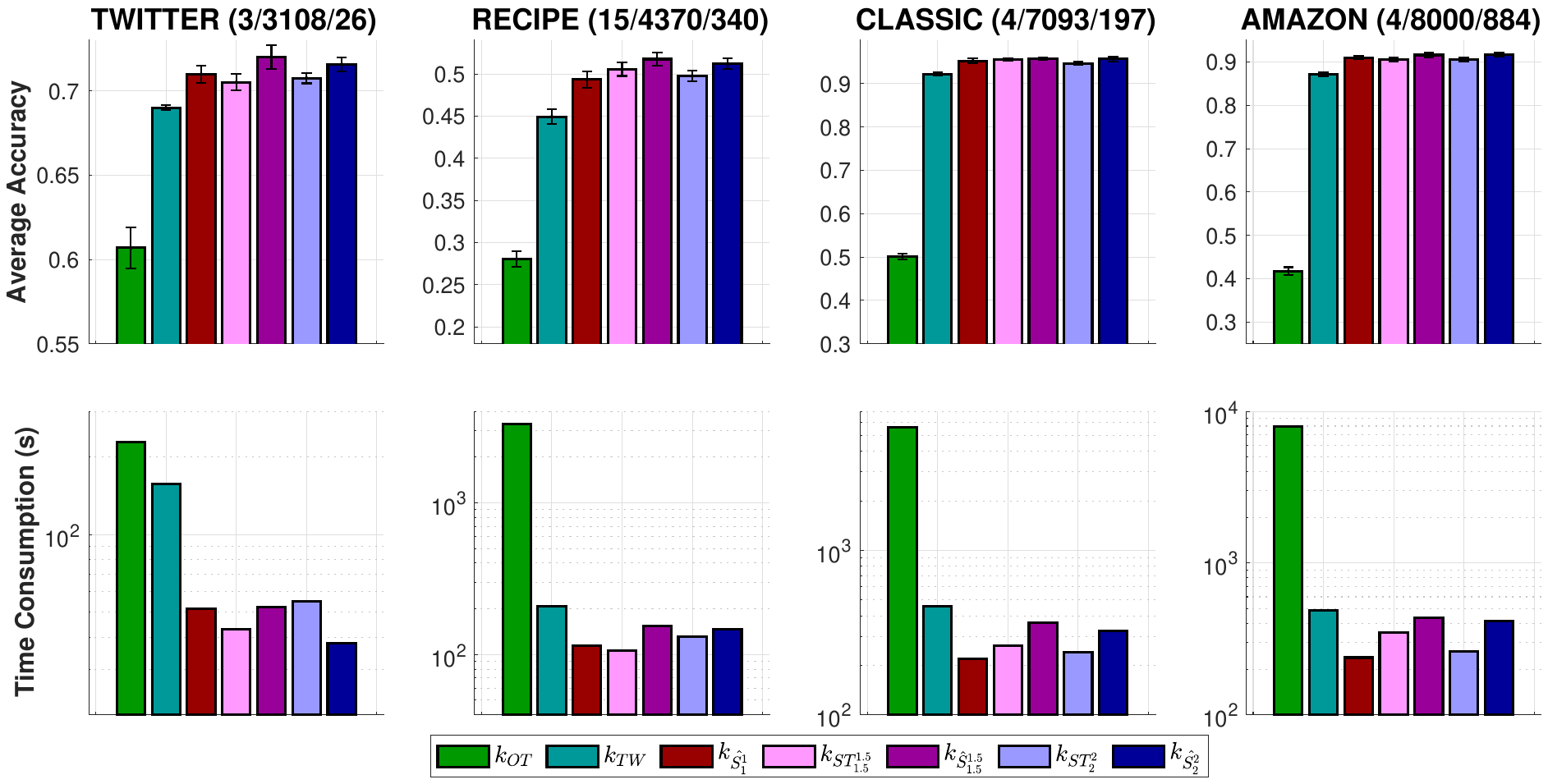}
  \end{center}
  \vspace{-6pt}
  \caption{SVM results and time consumption for kernel matrices in document classification with $\G_{\text{Sqrt}}$ and $M = 1000$.}
  \label{fg:DOC_1KSqrt_main}
 \vspace{-6pt}
\end{figure*}

\begin{figure*}[ht]
  \begin{center}
    \includegraphics[width=0.75\textwidth]{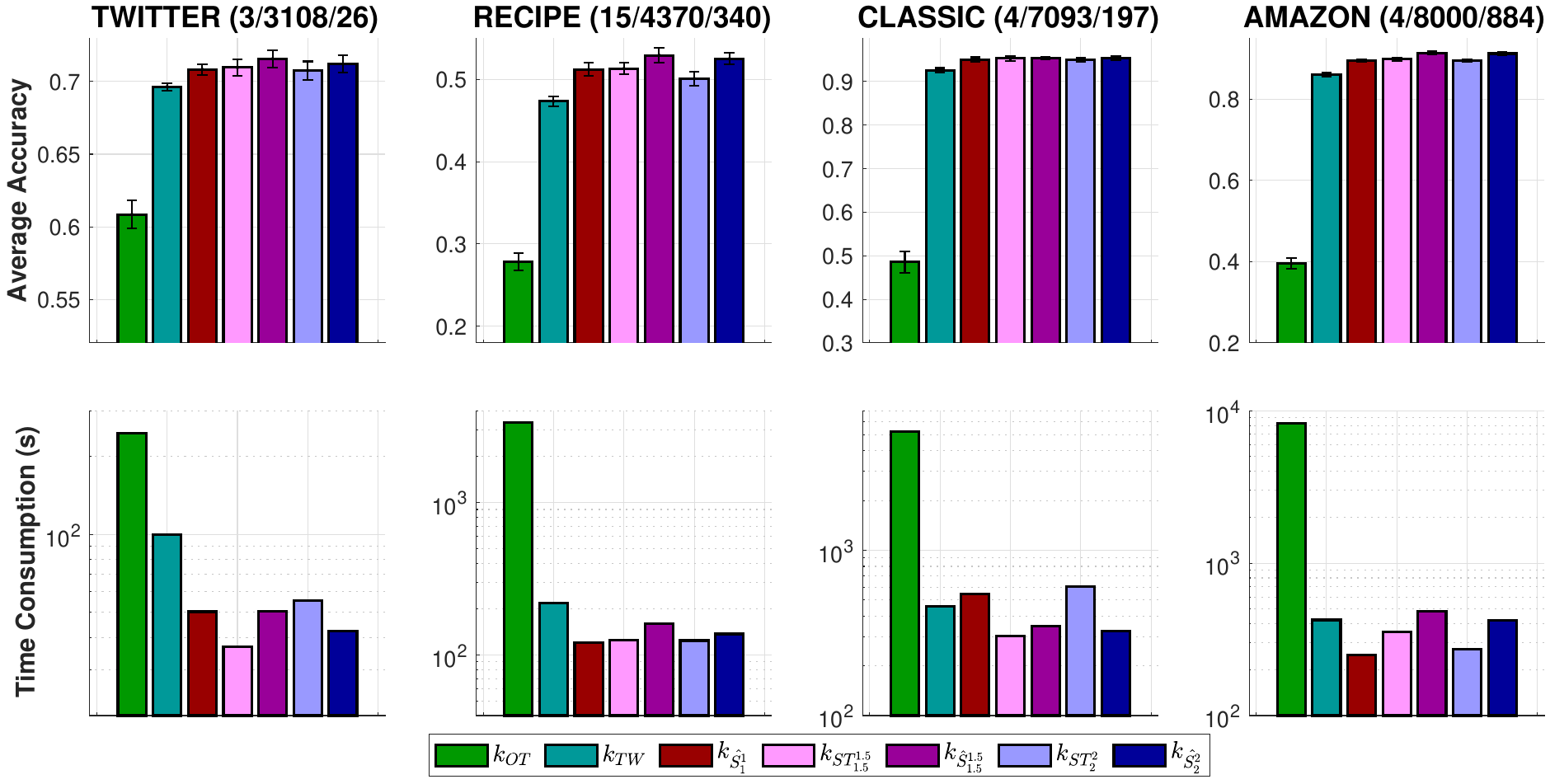}
  \end{center}
  \vspace{-6pt}
  \caption{SVM results and time consumption for kernel matrices in document classification with $\G_{\text{Log}}$ and $M = 1000$.}
  \label{fg:DOC_1KLog_main}
 \vspace{-6pt}
\end{figure*}

\begin{figure*}[ht]
  \begin{center}
    \includegraphics[width=0.75\textwidth]{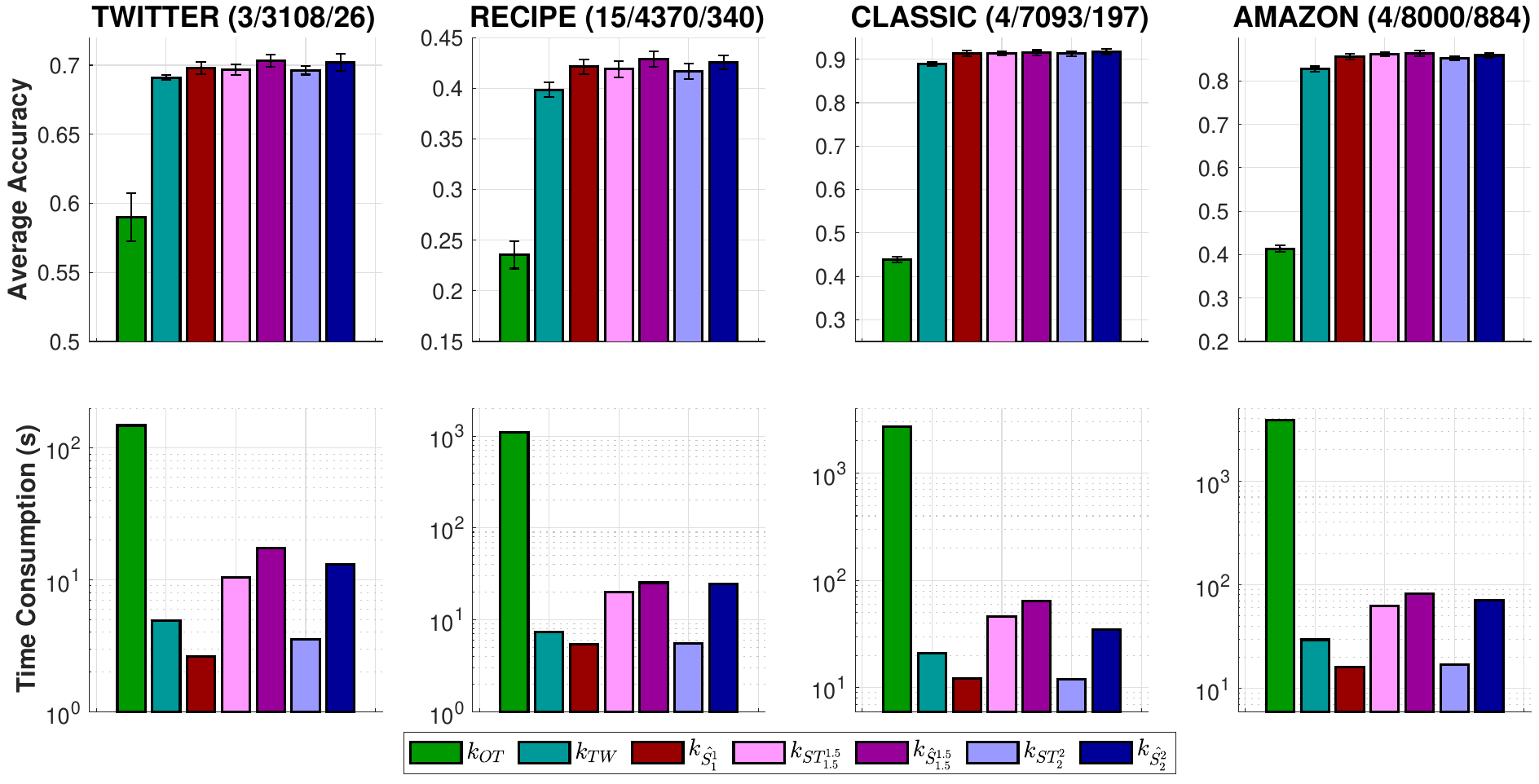}
  \end{center}
  \vspace{-6pt}
  \caption{SVM results and time consumption for kernel matrices in document classification with $\G_{\text{Sqrt}}$ and $M = 100$.}
  \label{fg:DOC_100Sqrt_main}
 \vspace{-6pt}
\end{figure*}

\begin{figure*}[ht]
  \begin{center}
    \includegraphics[width=0.75\textwidth]{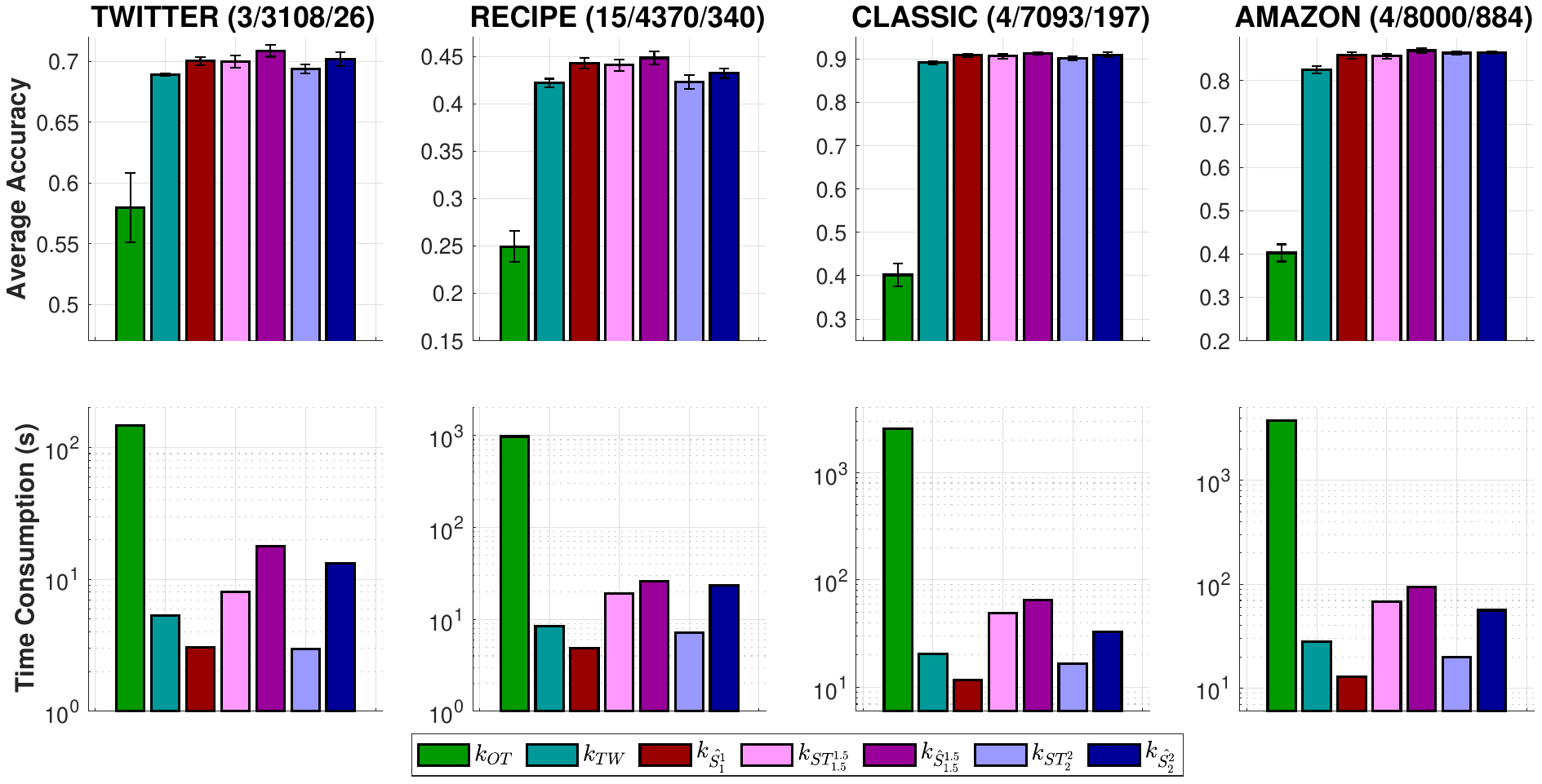}
  \end{center}
  \vspace{-6pt}
  \caption{SVM results and time consumption for kernel matrices in document classification with $\G_{\text{Log}}$ and $M = 100$.}
  \label{fg:DOC_100Log_main}
 \vspace{-6pt}
\end{figure*}

\paragraph{+ For document classification.}
\begin{itemize}
    \item Figures~\ref{fg:DOC_1KSqrt_main} and \ref{fg:DOC_1KLog_main} illustrate SVM results and time consumption for kernel matrices on graphs $\G_{\text{Sqrt}}$ and $\G_{\text{Log}}$ respectively where the number of nodes $M = 1000$.

    \item Figures~\ref{fg:DOC_100Sqrt_main} and \ref{fg:DOC_100Log_main} illustrate SVM results and time consumption for kernel matrices on graphs $\G_{\text{Sqrt}}$ and $\G_{\text{Log}}$ respectively where the number of nodes $M = 100$.
\end{itemize}

\begin{figure}[ht]
  \begin{center}
    \includegraphics[width=0.42\textwidth]{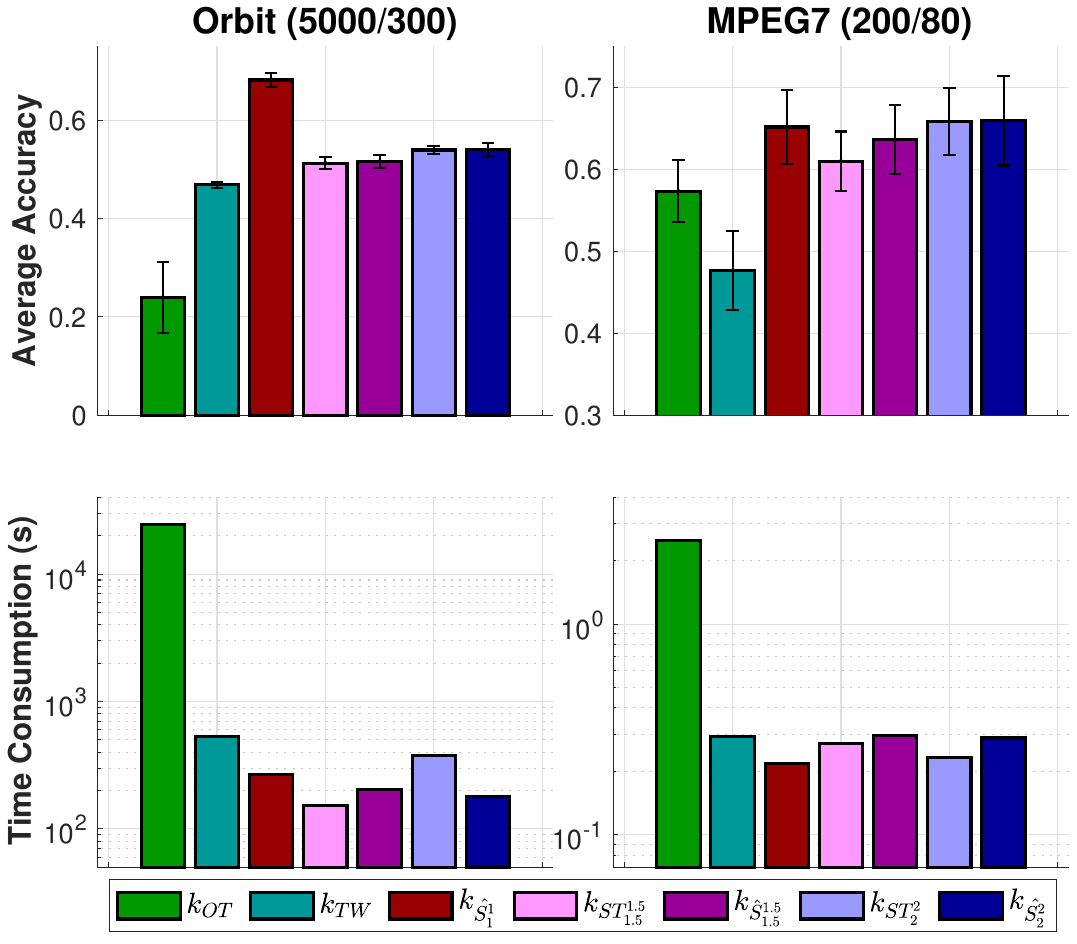}
  \end{center}
  \vspace{-6pt}
  \caption{SVM results and time consumption for kernel matrices in TDA with $\G_{\text{Sqrt}}$ and $M = 1000$.}
  \label{fg:TDA_1KSqrt_main}
 \vspace{-6pt}
\end{figure}

\begin{figure}[ht]
  \begin{center}
    \includegraphics[width=0.42\textwidth]{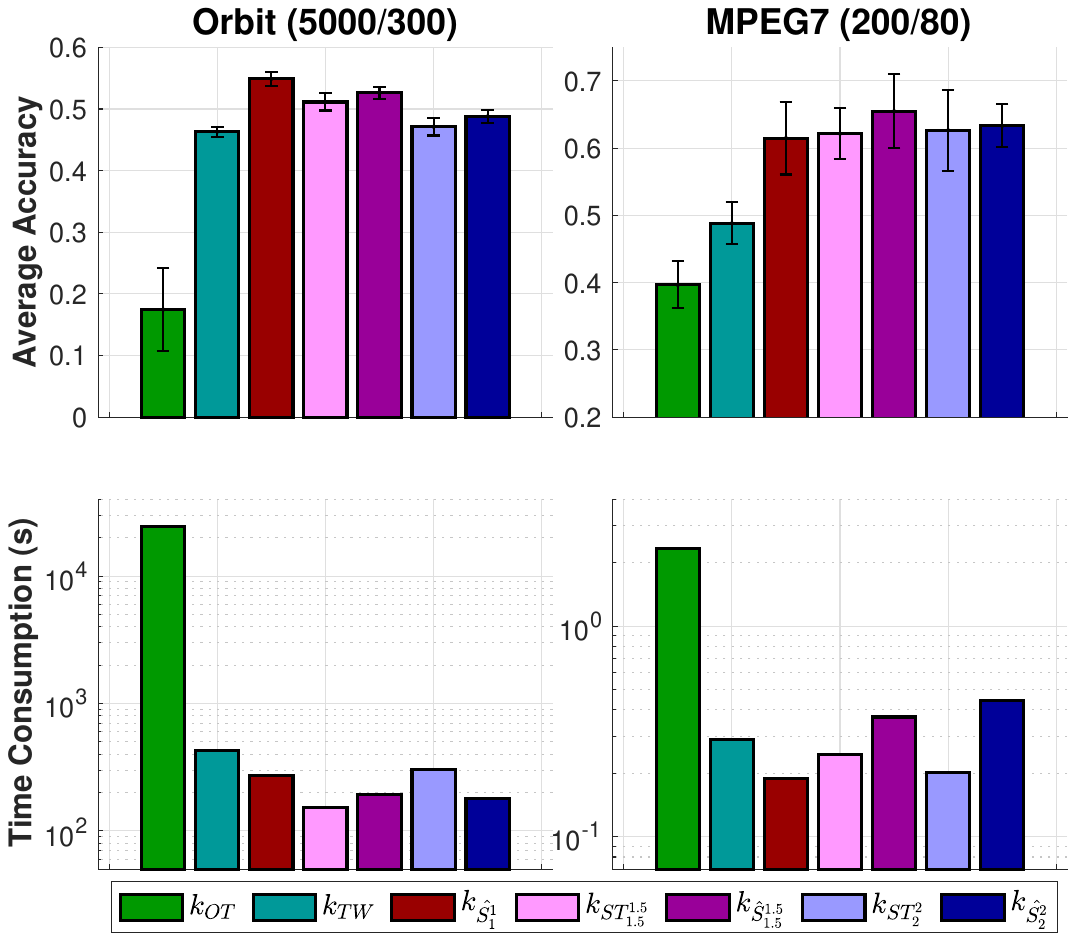}
  \end{center}
  \vspace{-6pt}
  \caption{SVM results and time consumption for kernel matrices in TDA with $\G_{\text{Log}}$ and $M = 1000$.}
  \label{fg:TDA_1KLog_main}
 \vspace{-6pt}
\end{figure}

\begin{figure}[ht]
  \begin{center}
    \includegraphics[width=0.42\textwidth]{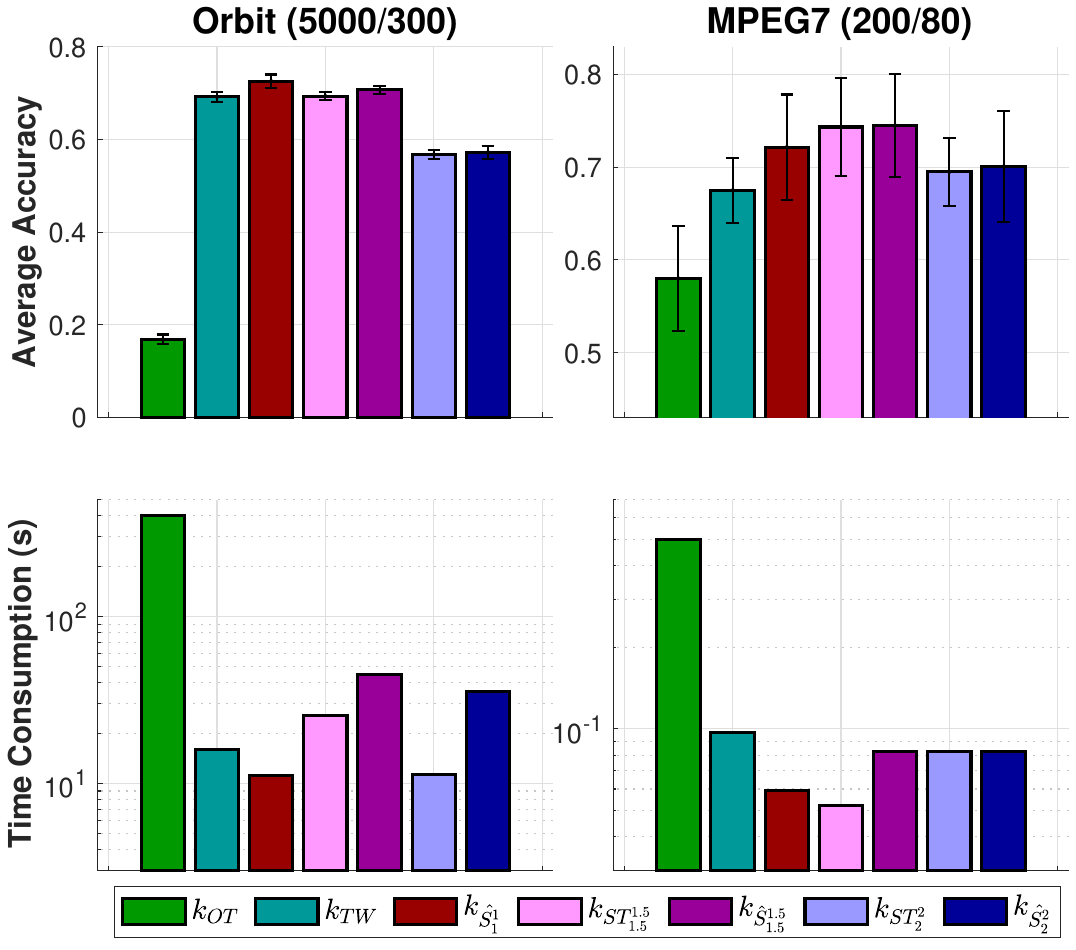}
  \end{center}
  \vspace{-6pt}
  \caption{SVM results and time consumption for kernel matrices in TDA with $\G_{\text{Sqrt}}$ and $M = 100$.}
  \label{fg:TDA_100Sqrt_main}
 \vspace{-6pt}
\end{figure}

\begin{figure}[ht]
  \begin{center}
    \includegraphics[width=0.42\textwidth]{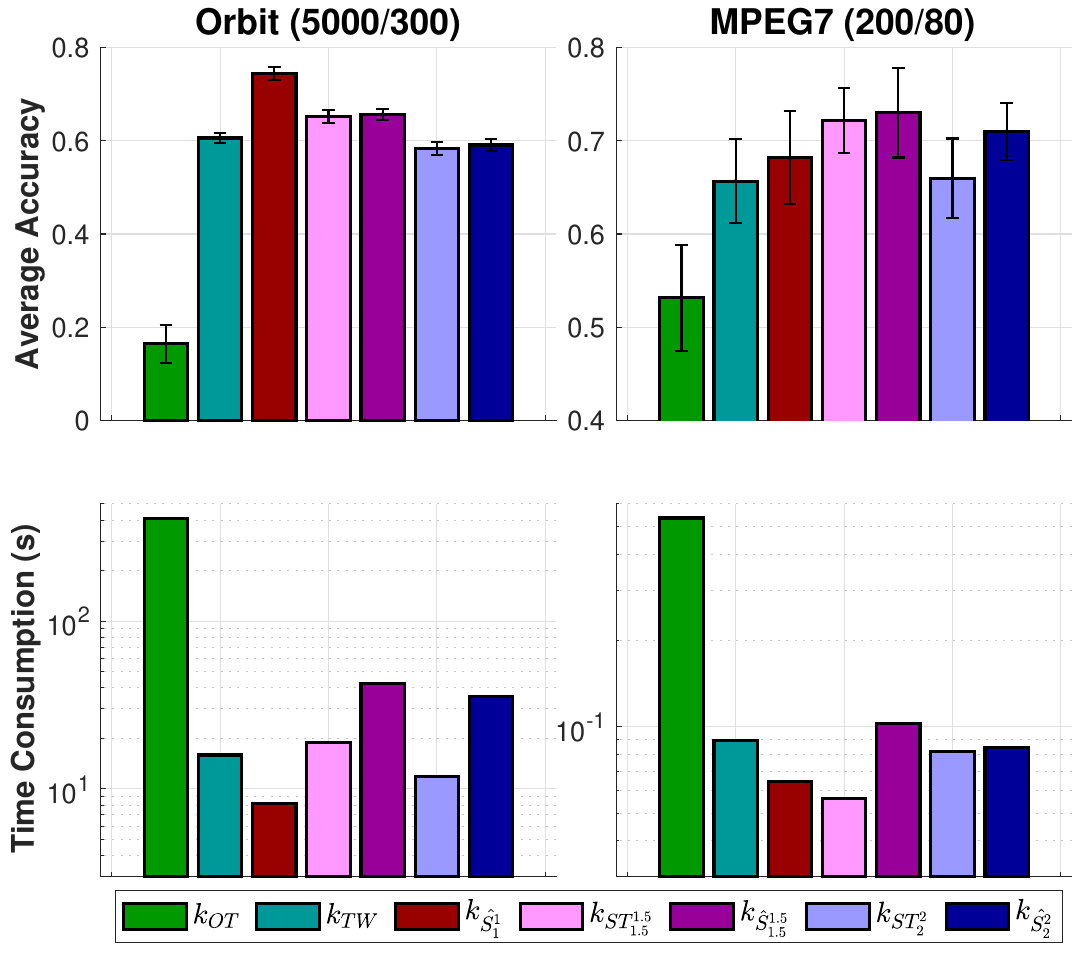}
  \end{center}
  \vspace{-6pt}
  \caption{SVM results and time consumption for kernel matrices in TDA with $\G_{\text{Log}}$ and $M = 100$.}
  \label{fg:TDA_100Log_main}
 \vspace{-6pt}
\end{figure}

\paragraph{+ For TDA.}
\begin{itemize}
    \item Figures~\ref{fg:TDA_1KSqrt_main} and \ref{fg:TDA_1KLog_main} illustrate SVM results and time consumption for kernel matrices on graphs $\G_{\text{Sqrt}}$ and $\G_{\text{Log}}$ respectively where the number of nodes $M = 1000$.

    \item Figures~\ref{fg:TDA_100Sqrt_main} and \ref{fg:TDA_100Log_main} illustrate SVM results and time consumption for kernel matrices on graphs $\G_{\text{Sqrt}}$ and $\G_{\text{Log}}$ respectively where the number of nodes $M = 100$.
\end{itemize}

We next provide further results for the regularized Sobolev IPM when varying the number of slices on both graphs $\G_{\text{Sqrt}}$ and $\G_{\text{Log}}$ for document classification and TDA.

\begin{figure*}[ht]
  \begin{center}
    \includegraphics[width=0.78\textwidth]{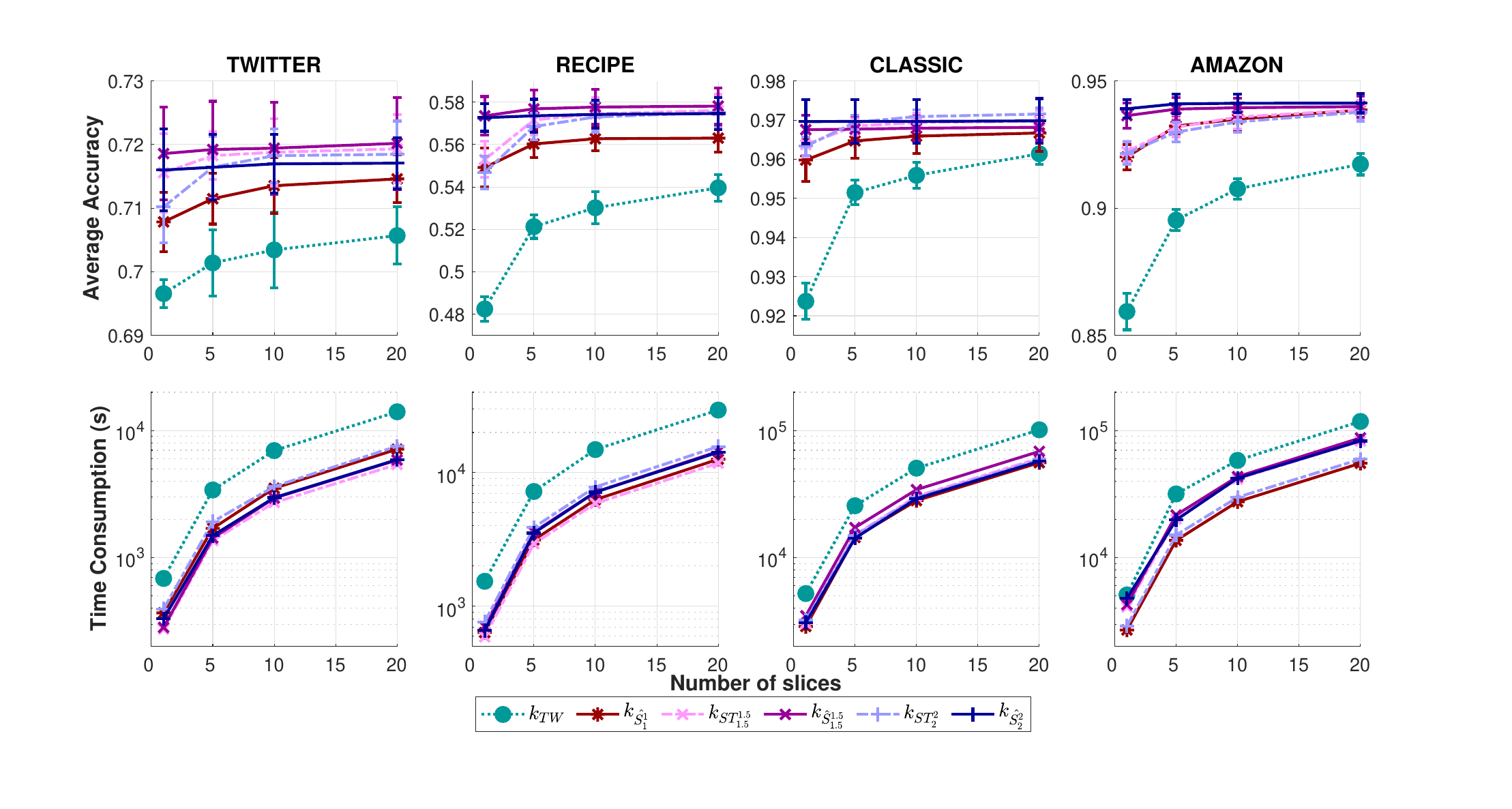}
  \end{center}
  \vspace{-6pt}
  \caption{SVM results and time consumption for kernel matrices of slice variants with $\G_{\text{Sqrt}}$ and $M = 10^4$.}
  \label{fg:DOC_10KPow_SLICE_main}
 \vspace{-6pt}
\end{figure*}

\begin{figure*}[ht]
  \begin{center}
    \includegraphics[width=0.78\textwidth]{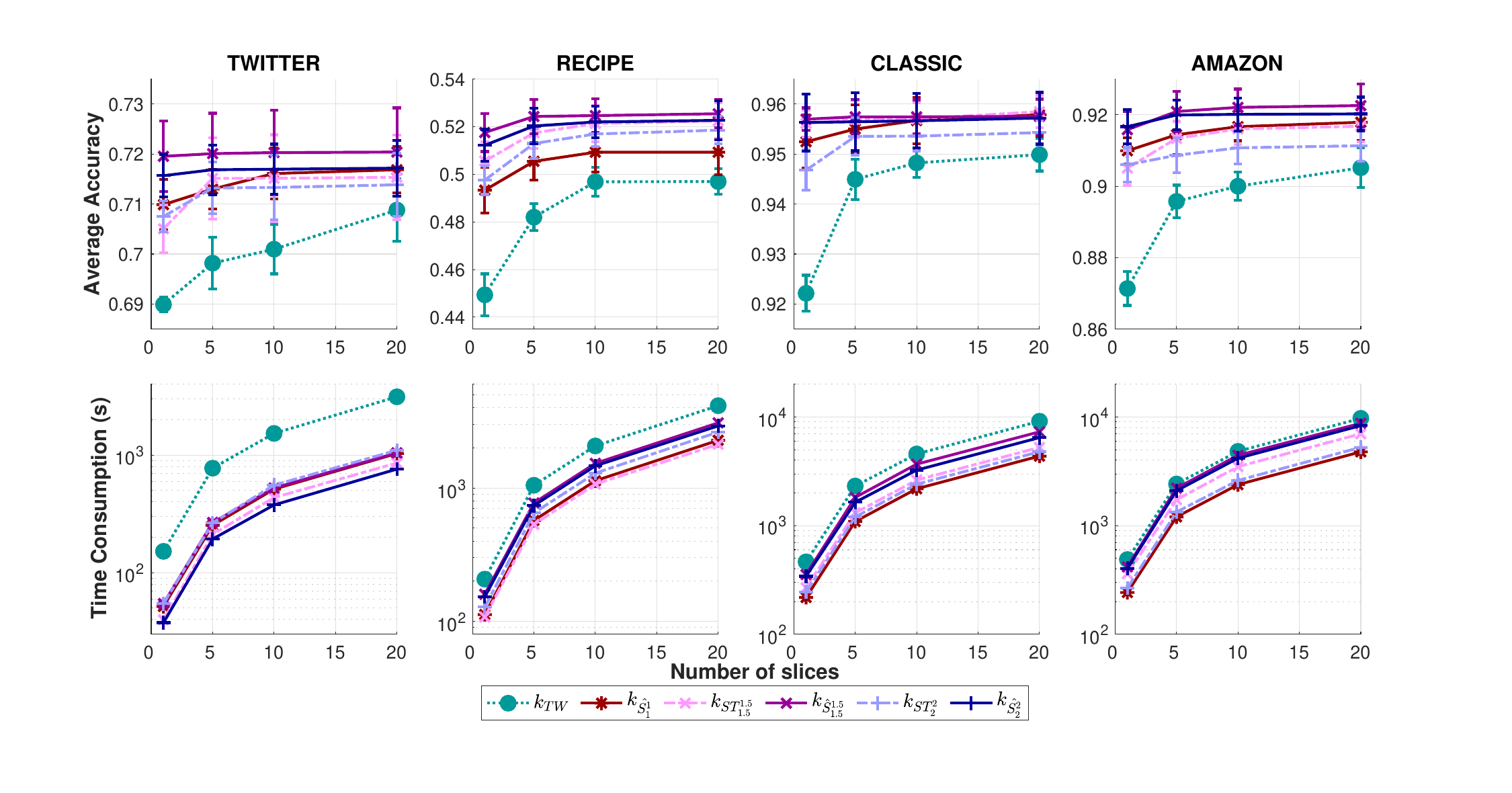}
  \end{center}
  \vspace{-6pt}
  \caption{SVM results and time consumption for kernel matrices of slice variants with $\G_{\text{Sqrt}}$ and $M = 1000$.}
  \label{fg:DOC_1KPow_SLICE_main}
 \vspace{-6pt}
\end{figure*}

\begin{figure*}[ht]
  \begin{center}
    \includegraphics[width=0.78\textwidth]{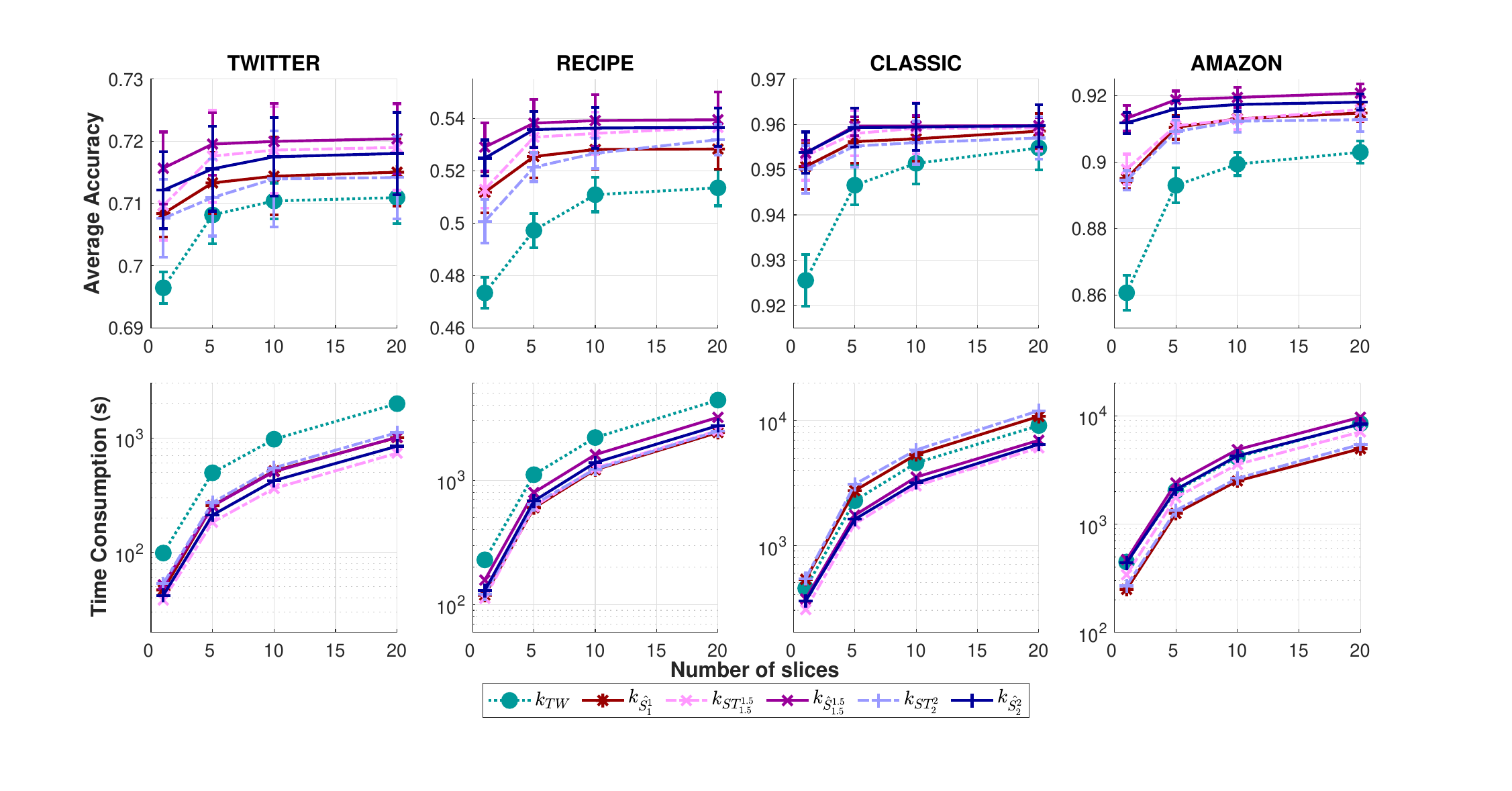}
  \end{center}
  \vspace{-6pt}
  \caption{SVM results and time consumption for kernel matrices of slice variants with $\G_{\text{Log}}$ and $M = 1000$.}
  \label{fg:DOC_1KLog_SLICE_main}
 \vspace{-6pt}
\end{figure*}

\begin{figure*}[ht]
  \begin{center}
    \includegraphics[width=0.78\textwidth]{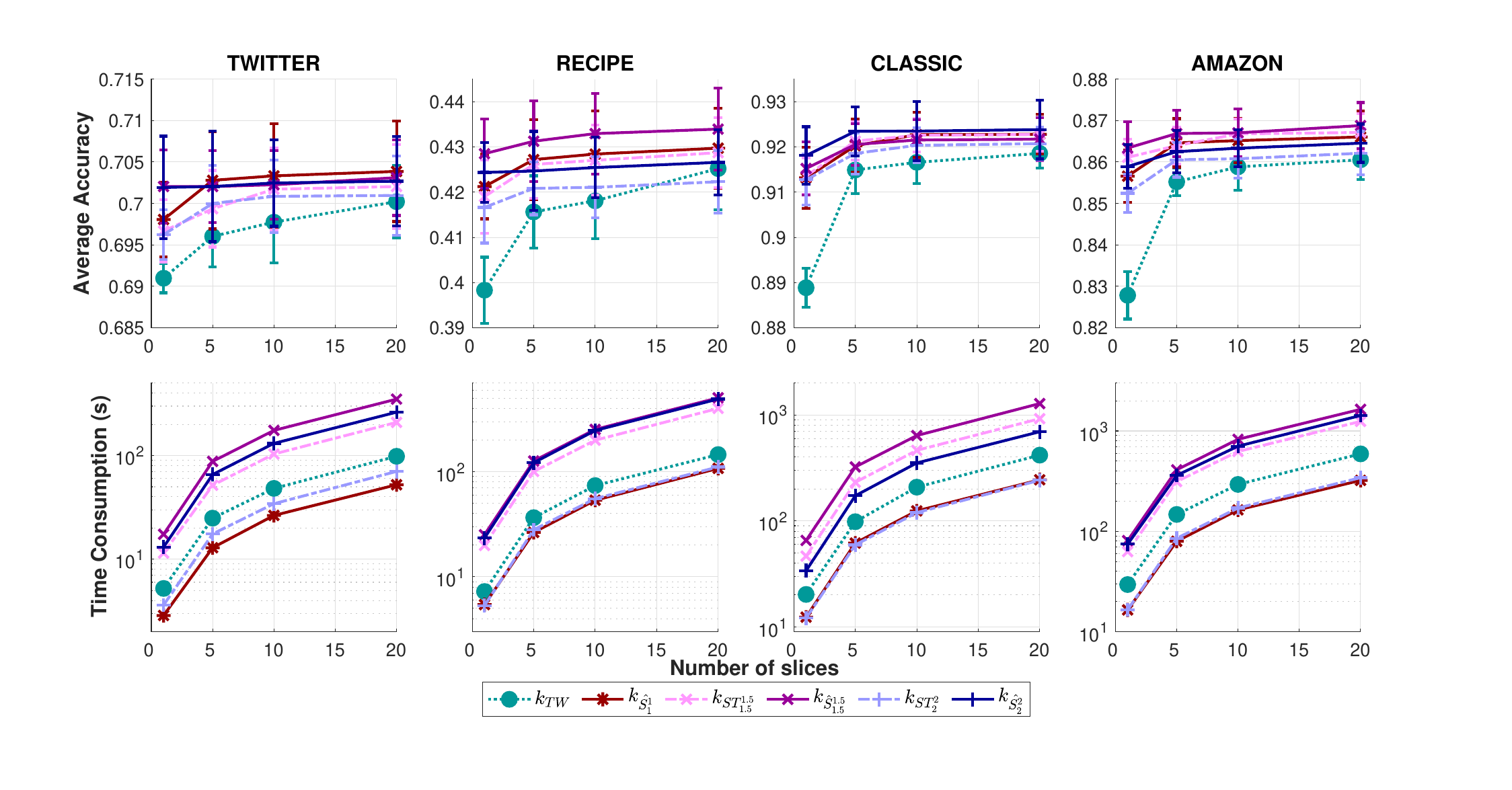}
  \end{center}
  \vspace{-6pt}
  \caption{SVM results and time consumption for kernel matrices of slice variants with $\G_{\text{Sqrt}}$ and $M = 100$.}
  \label{fg:DOC_100Pow_SLICE_main}
 \vspace{-6pt}
\end{figure*}

\begin{figure*}[ht]
  \begin{center}
    \includegraphics[width=0.78\textwidth]{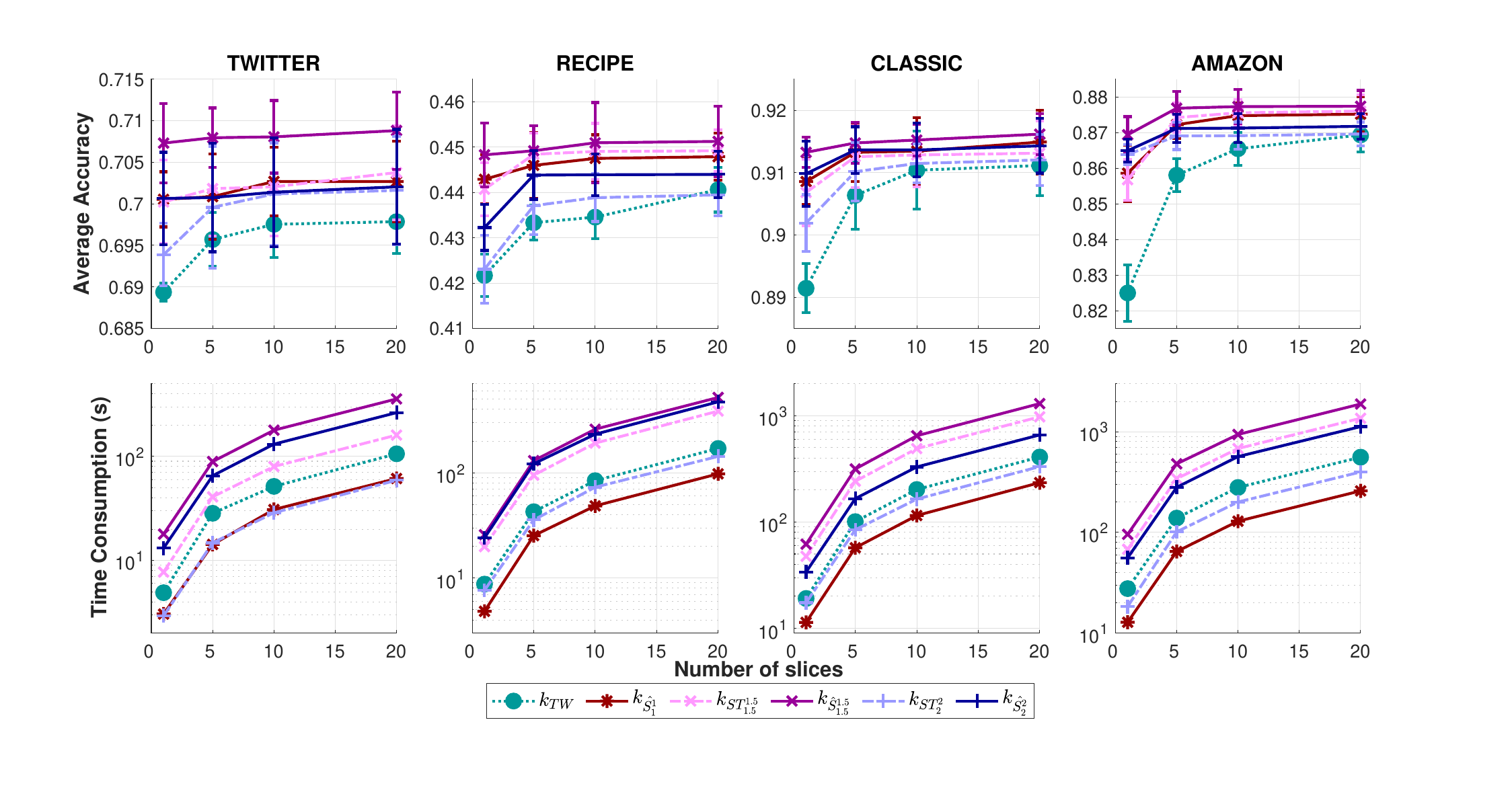}
  \end{center}
  \vspace{-6pt}
  \caption{SVM results and time consumption for kernel matrices of slice variants with $\G_{\text{Log}}$ and $M = 100$.}
  \label{fg:DOC_100Log_SLICE_main}
 \vspace{-6pt}
\end{figure*}

\paragraph{+ For document classification.}
\begin{itemize}
    \item Figure~\ref{fg:DOC_10KPow_SLICE_main} illustrates SVM results and time consumption for kernel matrices on graph $\G_{\text{Sqrt}}$ where the number of nodes $M = 10^4$.

    \item Figures~\ref{fg:DOC_1KPow_SLICE_main} and \ref{fg:DOC_1KLog_SLICE_main} illustrate SVM results and time consumption for kernel matrices on graphs $\G_{\text{Sqrt}}$ and $\G_{\text{Log}}$ respectively where the number of nodes $M = 1000$.

    \item Figures~\ref{fg:DOC_100Pow_SLICE_main} and \ref{fg:DOC_100Log_SLICE_main} illustrate SVM results and time consumption for kernel matrices on graphs $\G_{\text{Sqrt}}$ and $\G_{\text{Log}}$ respectively where the number of nodes $M = 100$.
\end{itemize}

\begin{figure}
  \begin{center}
    \includegraphics[width=0.42\textwidth]{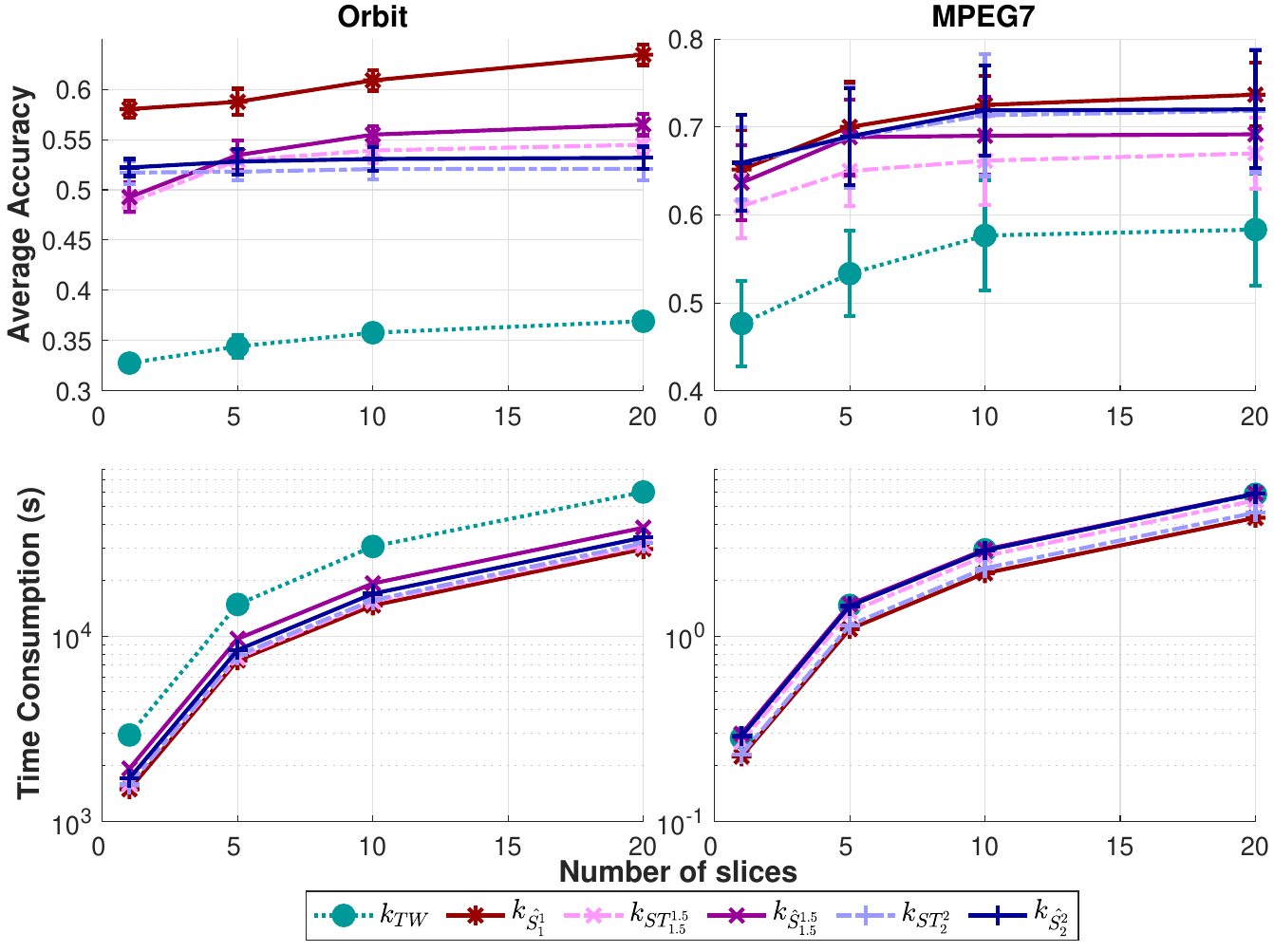}
  \end{center}
  \vspace{-6pt}
  \caption{SVM results and time consumption for kernel matrices of slice variants with graph $\G_{\text{Sqrt}}$ where the number of nodes $M = 10^4$ for \texttt{Orbit}, and $M=10^3$ for \texttt{MPEG7}.}
  \label{fg:TDA_mix10K1K_Pow_AccTime_SLICE}
 \vspace{-10pt}
\end{figure}

\begin{figure}
    \centering
\begin{subfigure}[b]{0.48\textwidth}
    \includegraphics[width=0.85\textwidth]{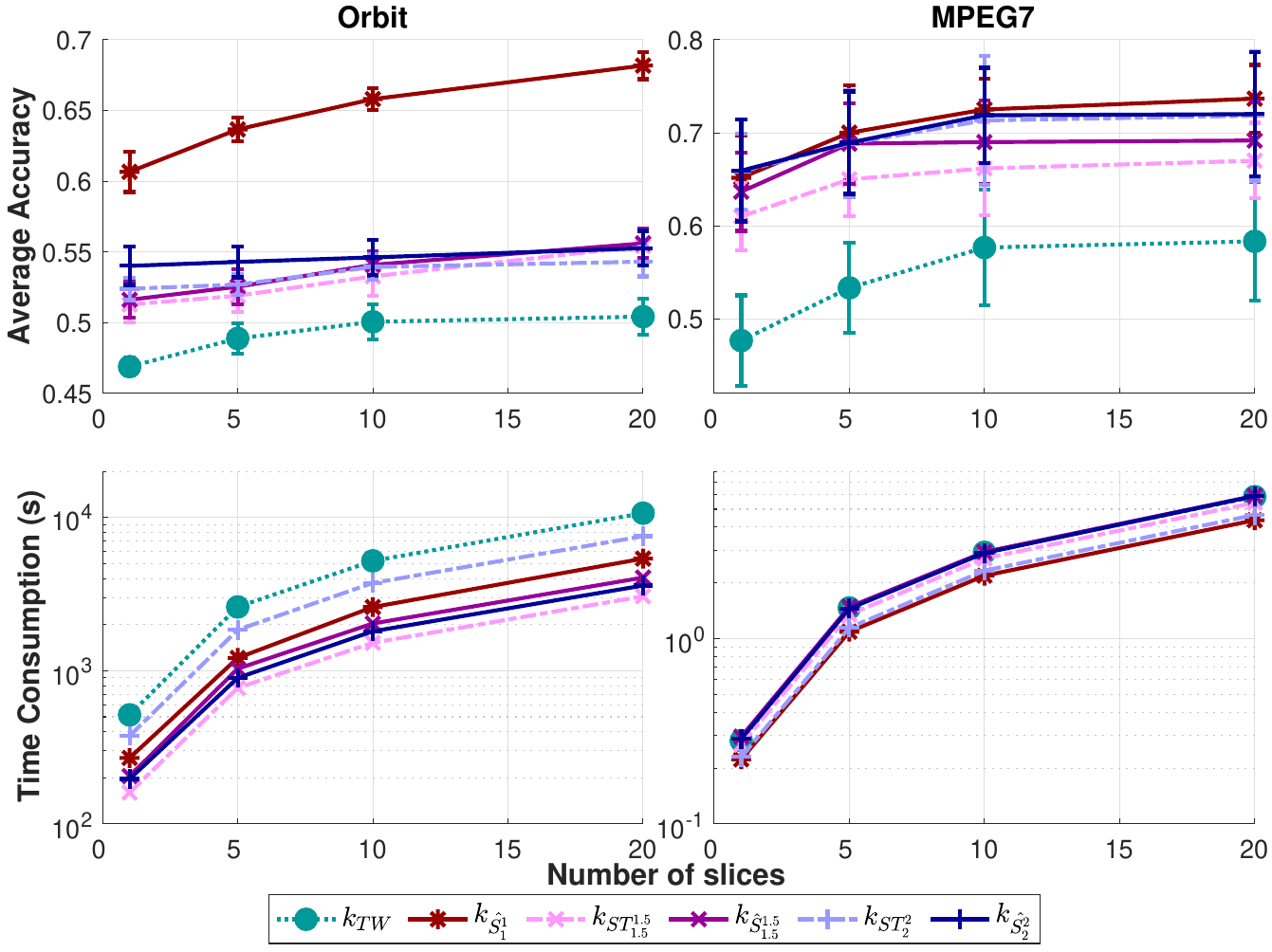}
  \caption{With $M=1000$.}
  \label{fg:TDA_1K_Pow_AccTime_SLICE}
 \vspace{-4pt}
\end{subfigure}
\hfill
\begin{subfigure}[b]{0.48\textwidth}
    \includegraphics[width=0.85\textwidth]{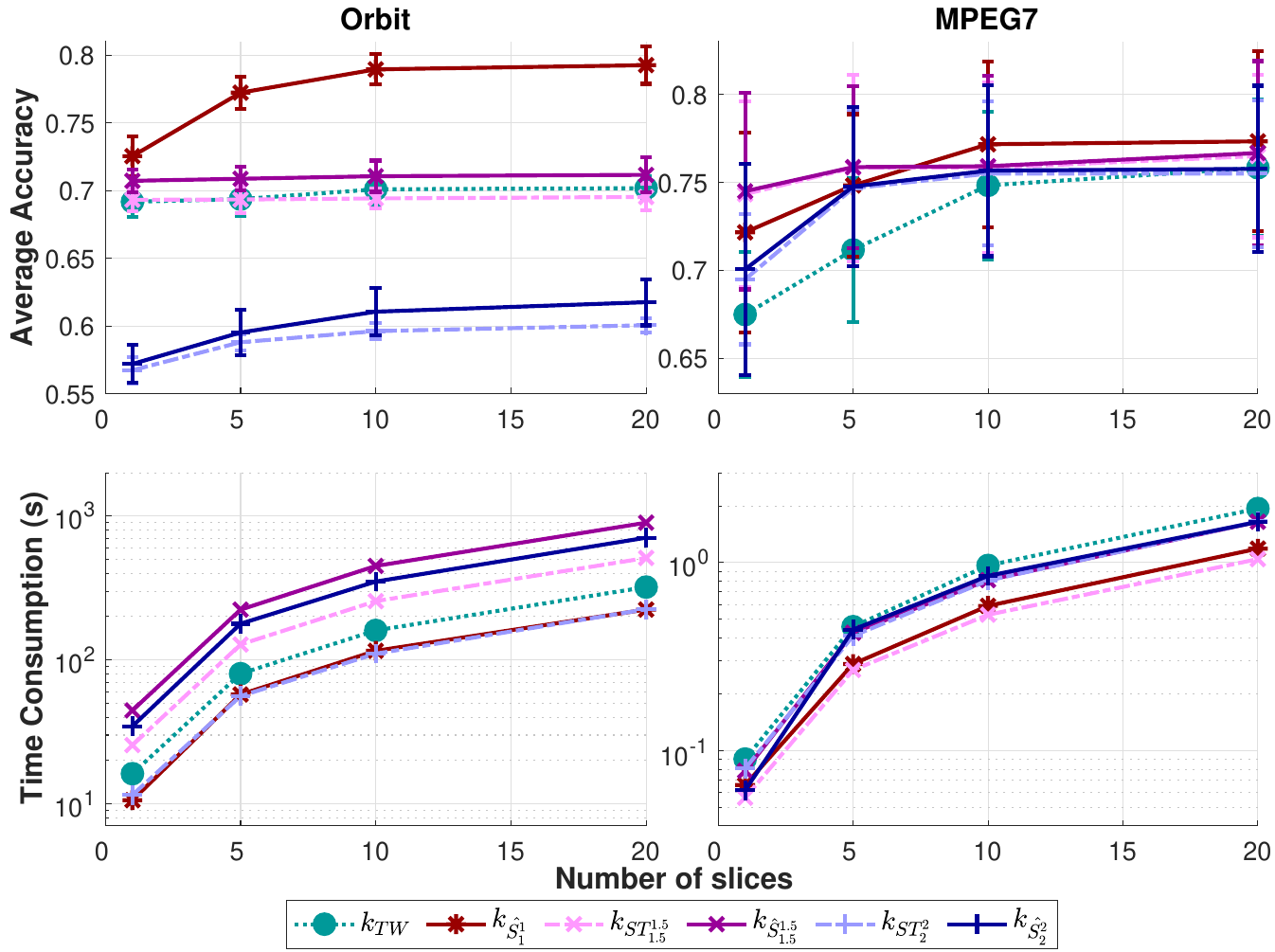}
  \caption{With $M=100$.}
  \label{fg:TDA_100_Pow_AccTime_SLICE}
 \vspace{-4pt}
\end{subfigure}
    \caption{SVM results and time consumption for kernel matrices of slice variants with graph $\G_{\text{Sqrt}}$.}
    \label{fg:Other_TDA_Pow_AccTime_SLICE}
\end{figure}

\begin{figure}
    \centering
\begin{subfigure}[b]{0.48\textwidth}
    \includegraphics[width=0.85\textwidth]{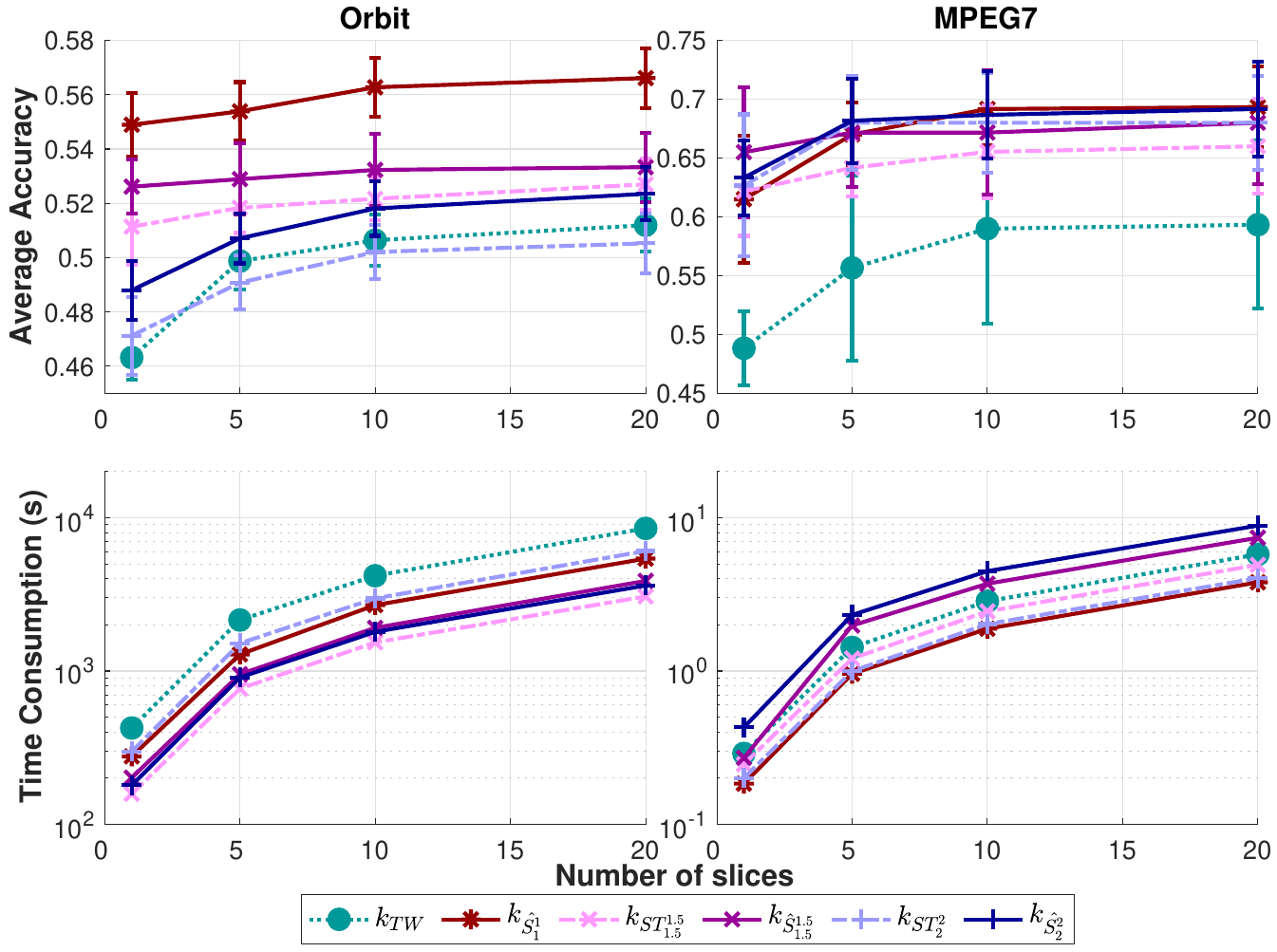}
  \caption{With $M=1000$.}
  \label{fg:TDA_1K_Log_AccTime_SLICE}
 \vspace{-4pt}
\end{subfigure}
\hfill
\begin{subfigure}[b]{0.48\textwidth}
    \includegraphics[width=0.85\textwidth]{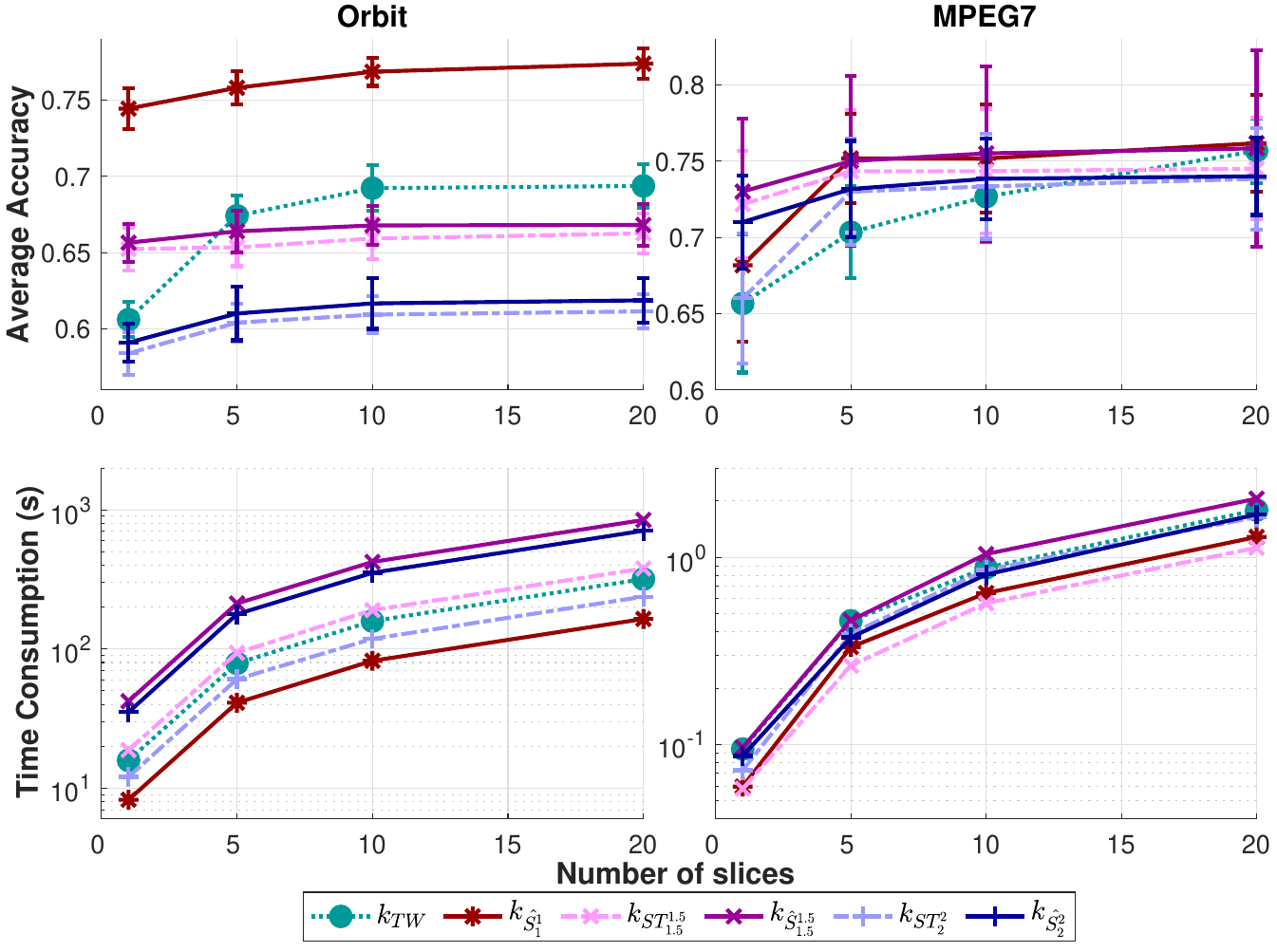}
  \caption{With $M=100$.}
  \label{fg:TDA_100_Log_AccTime_SLICE}
 \vspace{-4pt}
\end{subfigure}
    \caption{SVM results and time consumption for kernel matrices of slice variants with graph $\G_{\text{Log}}$.}
    \label{fg:Other_TDA_Log_AccTime_SLICE}
\end{figure}

\paragraph{+ For TDA.}
\begin{itemize}
    \item Figure~\ref{fg:TDA_mix10K1K_Pow_AccTime_SLICE} illustrates SVM results and time consumption for kernel matrices on graph $\G_{\text{Sqrt}}$ where the number of nodes $M = 10^4$ for \texttt{Orbit} and $M = 10^3$ for \texttt{MPEG7}.

    \item Figures~\ref{fg:TDA_1K_Pow_AccTime_SLICE} and \ref{fg:TDA_1K_Log_AccTime_SLICE} illustrate SVM results and time consumption for kernel matrices on graphs $\G_{\text{Sqrt}}$ and $\G_{\text{Log}}$ respectively where the number of nodes $M = 1000$.

    \item Figures~\ref{fg:TDA_100_Pow_AccTime_SLICE} and \ref{fg:TDA_100_Log_AccTime_SLICE} illustrate SVM results and time consumption for kernel matrices on graphs $\G_{\text{Sqrt}}$ and $\G_{\text{Log}}$ respectively where the number of nodes $M = 100$.
\end{itemize}

\subsection{Further Discussions}\label{app:sec:further_discussions}

For completeness, we recall important discussions on the underlying graph for Sobolev transport in~\citet{le2022st}, since they are also applied or adapted for the (regularized) Sobolev IPM in our work.

\paragraph{Measures on a graph.} In this work, we consider the (regularized) Sobolev IPM between \emph{two input probability measures} supported on the \emph{same} graph, which has the same setting as the Sobolev transport~\citep{le2022st}. 

The proposed kernels upon regularized Sobolev IPM are for \emph{input probability measures}, i.e., to compute similarity between two probability measures, on the \emph{same} graph. We distinguish our line of research to the following related problems: 

\textbf{$\bullet$ Compute distance between two (different) input graphs.} For examples,~\citet{petric2019got, dong2020copt, le2021fba, tam2022multiscale} compute OT problem (i.e., a specific instance of IPM) between \emph{two (different) input graphs}, where their goals are to compute distance between two input graphs. They are essentially different to our considered problem which computes distance (i.e., regularized Sobolev IPM) between \emph{two input probability measures} supported on the \emph{same} graph.

\textbf{$\bullet$ Graph kernels between two (different) input graphs.} Graph kernels are kernel functions between two input graphs to assess their similarity. A comprehensive review on graph kernels can be found in~\citet{borgwardt2020graph, kriege2020survey, nikolentzos2021graph}. Essentially, this line of research is different to our proposed kernels, i.e., regularized Sobolev IPM kernels to measures similarity between \emph{two input probability measures} on the \emph{same} graph.

\paragraph{Path length for points in graph $\G$~\citep{le2022st}.} We can canonically measure a path length connecting any two points  $x, y \in \G$ where these two points $x, y$ are not necessary to be nodes in $V$ of graph $\G$. 

Consider the edge $e= \langle u, v\rangle$ connecting two nodes $u, v \in V$, for $x, y \in \R^n$ and $x, y \in e$, we have 
\begin{eqnarray*}
& x = (1-s) u + s v, \\
& y = (1-t)u + t v,
\end{eqnarray*}
for some scalars $t,s\in [0,1]$. Thus, the length of the path connecting $x, y$ along the edge $e$ (i.e., the line segment $\langle x, y\rangle$) is equal to $|t-s| w_e$. Consequently, the length for an arbitrary path in $\G$ can be similarly defined by breaking down into pieces over edges and summing over their corresponding lengths~\citep{le2022st}.

\paragraph{Extension to measures supported on $\G$.} Much as Sobolev transport~\citep{le2022st}, the discrete case of the regularzed Sobolev IPM in Equation~\eqref{eq:regSobolevIPM_discrete} can be easily extended for measures with finite supports on $\G$ (i.e., supports of the input measures may not be nodes in $V$, but possibly points on edges in $E$) by using the same strategy to measure a path length for support data points in graph $\G$. More precisely, we break down edges containing supports into pieces and sum over their corresponding values instead of the sum over edges.

\paragraph{About the assumption of uniqueness property of the shortest paths on $\G$.} As discussed in \citet{le2022st} for Sobolev transport, note that $w_e \in \R_{+}$ for any edge $e \in E$ in graph $\G$., it is almost surely that every node in $V$ can be regarded as unique-path root node since with a high probability, lengths of paths connecting any two nodes in graph $\G$ are different.

Additionally, for some special graph, e.g., a grid of nodes, there is \emph{no} unique-path root node for such graph. However, by perturbing each node, and/or perturbing lengths of edges if $\G$ is a non-physical graph, with a small deviation, we can obtain a graph satisfying the unique-path root node assumption.

Besides that, for input probability measures with full supports in graph $\G$, or at least full supports in any cycle in graph $\G$, then it exists a special support data point where there are multiple shortest paths from the root node to it. In this case, we simply choose one fixed shortest path among them for this support data point (or we can add a virtual edge from the root node to this support data point where the edge length is deducted by a small deviation). In many practical applications (e.g., document classification and TDA in our experiments), one can neglect this special case since input probability measures have a finite number of supports.

\paragraph{About the regularized Sobolev IPM.} Much as Sobolev transport~\citep{le2022st}, we assume that the graph metric space (i.e., the graph structure) is given. We leave the question to learn an optimal graph metric structure adaptively from data for future work.

\paragraph{About the graphs $\G_{\text{Log}}$ and $\G_{\text{Sqrt}}$~\citep{le2022st}.} For a fast computation, we use the farthest-point clustering method to partition supports of measures into at most $M$ clusters.\footnote{$M$ is the input number of clusters used for the clustering method. Thus, we obtain at most $M$ clusters, depending on input data points.} Then, let the set of vertices $V$ be the set of centroids of these clusters, i.e., graph vertices. For edges, in graph $\G_{\text{Log}}$, we randomly choose $(M\log{M})$ edges; and $M^{3/2}$ edges for graph $\G_{\text{Sqrt}}$. We further denote the set of those randomly sampled edges as $\tilde{E}$.  

For each edge $e$, its corresponding edge length (i.e., edge weight) $w_e$ is computed by the Euclidean distance between the two corresponding nodes of edge $e$. Let $n_c$ be the number of connected components in the graph $\tilde{\G}(V, \tilde{E})$. Then, we randomly add $(n_c - 1)$ more edges between these $n_c$ connected components to construct a connected graph $\G$ from $\tilde{\G}$. Let $E_c$ be the set of these $(n_c - 1)$ added edges and denote set $E = \tilde{E} \cup E_c$, then $\G(V, E)$ is the constructed graph. 

\paragraph{Datasets and Computational Devices.} For the datasets in our experiments (i.e., \texttt{TWITTER, RECIPE, CLASSIC, AMAZON} for document datasets, and \texttt{Orbit, MPEG7} for TDA), one can contact the authors of Sobolev transport~\citep{le2022st} to access to them. For computational devices, we run all of our experiments on commodity hardware.

\paragraph{Hyperparamter validation.} For validation, we further randomly split \emph{the training set} into $70\%/30\%$ for validation-training and validation with $10$ repeats to choose hyper-parameters in our experiments.

\paragraph{The number of pairs in training and test for kernel SVM~\citep{le2024generalized}.} Let $N_{tr}, N_{te}$ be the number of measures used for training and test respectively. For the kernel SVM training, the number of pairs which we compute the distances is $(N_{tr}-1) \times \frac{N_{tr}}{2}$. For the test phase, the number of pairs which we compute the distances is $N_{tr} \times N_{te}$. Therefore, for each run, the number of pairs which we compute the distances for both training and test is totally $N_{tr} \times (\frac{N_{tr}-1}{2} +  N_{te})$. For examples, in Table~\ref{tb:numpairs}, we list these number of pairs for kernel SVM for all datasets used in our experiments.

\paragraph{Distinguish to the approach in~\citet{peyre2018comparison}.} We study the Sobolev IPM for probability measures supported on a given graph. The Sobolev IPM (Equation~\eqref{eq:SobolevIPM}) constrains a critic function within the unit ball defined by the Sobolev norm (Equation~\eqref{eq:SobolevNorm}) involving both the critic function and its gradient, while~\citet{peyre2018comparison} considers the \emph{weighted homogeneous Sobolev norm} which constraints a critic function in a \emph{semi-norm involving only gradient of the critic function}~\citet[Equations (3) and (4)]{peyre2018comparison}.~\citet{peyre2018comparison}'s approach shares the same spirit as the 
$2$-order Sobolev transport~\citep{le2022st}.

The proposed regularized Sobolev IPM may share some spirit with the weighted homogeneous Sobolev norm and the Sobolev transport (i.e., constraint on gradient of a critic function). However, we clarify that the weight function (Equation~\eqref{eq:weighting_func_wLp}) plays the key role to establish the equivalence between Sobolev norm and weighted norm (Theorem~\ref{thrm:S_wLp_norm}) for our novel regularization for Sobolev IPM.

\paragraph{Further discussions on the research problem and approach.} We emphasize that our approach aims neither to derive a \emph{tight bound} for the gap nor to provide a \emph{sharp approximation} for the challenging optimization problem (Equation~\eqref{eq:SobolevIPM}) of the original Sobolev IPM. Our purpose is in a different direction. More precisely, we instead propose a \emph{novel regularization} for Sobolev IPM which yields an \emph{equivalent metric} to the original Sobolev IPM, and a \emph{closed-form expression for a fast computation}. Note that, to our knowledge, there are \emph{no efficient algorithms} to compute Sobolev IPM effectively. We believe that deriving tight bound or providing sharp approximation for the Sobolev IPM is interesting research direction, and is left for future investigation.


\end{document}